\documentclass[runningheads]{llncs}

\usepackage{eccv}

\usepackage{eccvabbrv}

\usepackage{graphicx}
\usepackage{booktabs}
\usepackage[table]{xcolor}

\usepackage[accsupp]{axessibility}  

\usepackage{booktabs}
\usepackage{array}
\usepackage{tabularx}
\usepackage{stfloats}
\usepackage{caption}

\usepackage[hidelinks,breaklinks,colorlinks,citecolor=eccvblue]{hyperref}

\usepackage{orcidlink}
\usepackage{bibunits}

\usepackage{enumitem}

\usepackage{soul}

\sethlcolor{cyan!18}
\newcommand{\firstterm}[1]{\hl{#1}}
\newcommand{\reuseterm}[1]{\ul{#1}}

\makeatletter
\let\tf@llncs@section\section
\let\tf@llncs@subsection\subsection
\let\tf@llncs@subsubsection\subsubsection
\let\tf@llncs@subparagraph\subparagraph
\let\subparagraph\paragraph
\makeatother
\usepackage{titlesec}
\makeatletter
\let\subparagraph\tf@llncs@subparagraph
\makeatother
\titlespacing*{\section}{0pt}{6pt plus 0pt minus 0pt}{6pt plus 0pt minus 0pt}
\titlespacing*{\subsection}{0pt}{4pt plus 0pt minus 1pt}{4pt plus 0pt minus 1pt}
\titlespacing*{\subsubsection}{0pt}{3pt plus 0pt minus 1pt}{3pt plus 0pt minus 1pt}

\begin{document}

\setlist[itemize]{label=\textbullet, topsep=0.3\baselineskip, itemsep=0.15\baselineskip, parsep=0pt, partopsep=0pt}

\setlength{\abovedisplayskip}{3pt plus 0.5pt minus 0.5pt}
\setlength{\belowdisplayskip}{3pt plus 0.5pt minus 0.5pt}
\setlength{\abovedisplayshortskip}{0.5pt plus 0.5pt}
\setlength{\belowdisplayshortskip}{0.5pt plus 0.5pt}

\title{Where Does Generative Difficulty Reside? \\An Empirical Study of Target Representations} 

\titlerunning{On Generative Difficulty}

\author{
Marcel Plocher\inst{1}\orcidlink{0009-0008-8112-9610} \and \\
Bernhard Sch\"olkopf\inst{2,3,5}\orcidlink{0000-0002-8177-0925} \and 
Andreas Geiger\inst{1,4,5}\orcidlink{0000-0002-8151-3726} \and \\
Gege Gao\inst{1,2}\orcidlink{0000-0002-6918-2928}\textsuperscript{\textdagger}
}

\authorrunning{M. Plocher et al.}

\institute{
University of T\"ubingen, T\"ubingen, Germany \\
\email{marcel.plocher@student.uni-tuebingen.de} \vspace{0.4em} \and
ETH Z\"urich, Z\"urich, Switzerland \\ 
\email{\{gege.gao,a.geiger\}@uni-tuebingen.de} \vspace{0.4em} \and
Max Planck Institute for Intelligent Systems, T\"ubingen, Germany 
\email{bs@tuebingen.mpg.de} \vspace{0.4em}
\and
T\"ubingen AI Center, T\"ubingen, Germany  
\and
ELLIS Institute, T\"ubingen, Germany
\\[0.8em] 
\textsuperscript{\textdagger} \textbf{Project lead.}  
}

\maketitle

\begin{abstract}
  The target representation defines the distribution an image generator must learn, yet it is often treated as an interchangeable interface. This assumption is particularly questionable for continuous masked generators, which combine contextual inference from visible tokens with conditional modeling of each missing token. We study raw pixels, SD-VAE latents and DINOv2 as well as MAE representation-autoencoder features within a unified masked autoregressive rectified-flow model. Under a shared ImageNet training budget, these spaces exhibit distinct optimization and inference regimes. DINOv2 converges fastest in both iterations and computation but benefits strongly from a wider local denoiser and direct context fusion. Pixels optimize substantially more slowly and require a different prediction, masking, and guidance configuration. MAE reconstructs images more faithfully and exhibits clear semantic clustering, yet produces generations substantially worse than DINOv2. The representations also respond differently to classifier-free guidance and occupy distinct precision-recall trade-offs. Together, our results show that compression, reconstruction fidelity, token dimensionality, and visible semantic clustering do not individually predict generative behavior. Instead, target representations redistribute difficulty across contextual modeling, per-token denoising, and inference-time distributional control.

  \keywords{Autoregressive Generation \and Target Representations \and Rectified Flow}
\end{abstract}

\section{Introduction}
\label{sec:intro}
The representation in which a generative model predicts its targets determines more than the dimensionality of its output. It determines what information has already been removed or abstracted, how spatially localized each token is, how strongly tokens depend on their context, and what conditional distribution remains to be learned. Nevertheless, target representations are often treated as interchangeable interfaces: once an encoder and decoder are available, a generative architecture is expected to transfer across spaces with limited modification. We investigate when this assumption fails.

Most continuous image generators operate in the latent space of a variational autoencoder trained for image reconstruction. In particular, the Stable Diffusion VAE (SD-VAE) combines spatial and channel compression with sufficient perceptual reconstruction quality, substantially reducing the cost of high-resolution generation \cite{rombachHighResolutionImageSynthesis2022}. More recently, representation autoencoders have enabled generation in features from pretrained vision encoders such as DINOv2 and MAE \cite{oquabDINOv2LearningRobust2024,heMaskedAutoencodersAre2022,zhengDiffusionTransformersRepresentation2026}. Joint diffusion and flow models trained within these spaces can converge rapidly. Complementary work shows that direct pixel generation can also be effective under suitable prediction and architectural choices \cite{liBackBasicsLet2026}. These results establish the target representation as an important determinant of generative optimization.

Their implications for masked generation, however, remain unclear. Masked autoregressive modeling provides a particularly informative testbed for studying this question: by decomposing image generation into conditional token predictions, it exposes how the target representation shapes both contextual inference and the local prediction problem \cite{liAutoregressiveImageGeneration2024,youEffectiveEfficientMasked2025}. In continuous masked generators, a transformer aggregates visible tokens into contextual features, while a local diffusion or flow network models the conditional distribution of each missing token. A representation may therefore simplify one component while placing greater demands on the other.

This decomposition motivates our central question: \textbf{where does generative difficulty reside for different target representations?} Several intuitive explanations are possible. Compressed spaces may be easier because fewer values must be predicted. Reconstruction-oriented representations may be preferable because they preserve image information faithfully. Semantic representations may provide more informative and globally organized tokens. These properties are strongly entangled in existing systems, and it is unclear whether any of them alone predicts masked generative behavior.

We study this question using a unified masked autoregressive generator with a per-token rectified-flow objective. The model separates a transformer-based context network from a local conditional denoiser and supports four target spaces with the same 16$\times$16 token layout: raw image patches, compressed SD-VAE latents, self-distilled DINOv2 features, and reconstruction-trained MAE features. We train class-conditional ImageNet generators at 256$\times$256 resolution under a shared 503k-step (200 epochs) budget and compare optimization in both iterations and accumulated computation. Within each representation, we additionally examine masking, prediction parameterization, context-model and denoiser design, flow integration, classifier-free guidance, and precision–recall. 

We observe markedly different optimization regimes. Without classifier-free guidance, DINOv2 reaches an FID of 6.43, compared with 15.98 for SD-VAE and 43.60 for pixels, and maintains this advantage when training progress is measured in GFLOPs. However, its performance does not result from uniformly simplifying the architecture: DINOv2 performs well with a two-layer context encoder but benefits strongly from a wider local denoiser and direct context fusion. It also retains recognizable generations with a single Euler step, whereas SD-VAE and pixels degrade substantially under the same sampling procedure. Pixel generation remains viable, but benefits from clean-data prediction, a less aggressive masking distribution, a bottleneck, a stronger local denoiser, and linear classifier-free guidance.

The comparison between DINOv2 and MAE further challenges simple explanations based on representation properties. MAE provides better reconstruction quality, and both spaces exhibit visible semantic clustering, yet DINOv2 produces substantially better generations under the same generator configuration. Representation choice also changes inference-time behavior: DINOv2 obtains its best result without classifier-free guidance and combines high precision with low recall, whereas SD-VAE, MAE, and pixels benefit from representation-specific guidance strategies.

Together, these results show that compression, reconstruction fidelity, dimensionality, and visible semantic organization are individually insufficient to predict masked generative behavior. Instead, the target representation changes where difficulty appears: across contextual modeling, local conditional denoising, flow integration, and distributional control. 

Our contributions are threefold:
\begin{itemize}
    \item We instantiate masked autoregressive rectified flow across raw pixels, compressed VAE latents, and two representation-autoencoder feature spaces within a common generative factorization.
    \item We compare their optimization under a shared dataset, token layout, and training budget, and document the representation-specific modeling choices required by each space.
    \item We show that common representation properties do not individually predict generative behavior; instead, different target spaces induce distinct allocations of model capacity, sampling efficiency, guidance, and distribution coverage.
\end{itemize}

\section{Preliminaries}

Let an image be represented as a sequence
\begin{equation}
    x=\{x_1,\ldots,x_N\}, \quad x_i\in\mathbb{R}^{D},
\end{equation}
where $N=256$ in all experiments and $D$ depends on the target space. For latent representations, the sequence is produced by a frozen encoder and is mapped back to pixels by the corresponding decoder. For raw pixels, the image is patchified before generation and unpatchified afterward.

\subsection{Random-Order Autoregression. }
Given a random mask $M\subseteq\{1,\ldots,N\}$, the visible tokens are processed by a bidirectional transformer. The transformer produces a contextual vector $z_i$ for every masked location. Following random-order masked autoregression, the image distribution can be expressed through conditional predictions over subsets of tokens rather than through a fixed causal ordering:
\begin{equation}
    p(x)=\prod_k p(x_{M_k}\mid x_{\bar M_k}),
\end{equation}
where $M_k$ denotes the subset generated at autoregressive step $k$, and $\bar M_k$ denotes the tokens already visible at that step.

The transformer is therefore responsible for inter-token reasoning: it aggregates the visible image content and class-conditioning tokens into contextual features. The conditional distribution of each target token is modeled separately by a local flow network.

\subsection{Per-Token Rectified Flow}
For each masked token $x_i$, we independently sample Gaussian noise $\epsilon_i\sim\mathcal{N}(0,I)$ and a continuous timestep $t_i\in[0,1]$. We use the rectified-flow interpolation
\begin{equation}
    x_{t,i}=(1-t_i)x_i+t_i\epsilon_i,
\end{equation}
where $t_i=0$ corresponds to clean data and $t_i=1$ to Gaussian noise. The target velocity along this linear path is
\begin{equation}
    v_{t,i}=\epsilon_i-x_i.
\end{equation}

A local denoiser $v_\theta$ receives the noisy token $x_{t,i}$, its timestep $t_i$, and the contextual feature $z_i$. The complete model is trained with
\begin{equation}
    \mathcal{L}_\text{flow} = \mathbb{E}
    \left[
    \frac{1}{|M|D}
    \sum_{i\in M}
    \lVert
    v_\theta(x_{t,i},z_i,t_i)-v_{t,i}
    \rVert_2^2
    \right].
\end{equation}
The loss is computed only on masked positions and is normalized by both the number of masked tokens and the token dimension. Although the model expresses an autoregressive factorization, all masked tokens and their independently sampled timesteps are processed in parallel during training.

For the raw-pixel experiments, we additionally evaluate clean-data prediction (x-prediction). The denoiser directly predicts $x_i$, and the output is reparameterized as a velocity for flow integration
\begin{equation}
    v_\theta(x_{t,i},z_i,t_i) = \frac{x_\theta(x_{t,i},z_i,t_i)-x_{t,i}}{1-t_i}
\end{equation}
This intervention follows the motivation that clean image data may be easier for a capacity-limited network to predict than an off-manifold velocity target \cite{liBackBasicsLet2026}. Both prediction parameterizations are evaluated within the existing pixel and DINOv2 experiments.

\begin{figure}[t]
     \centering
     \begin{subfigure}[t]{.95\linewidth}
         \centering
         \includegraphics[width=.95\linewidth]{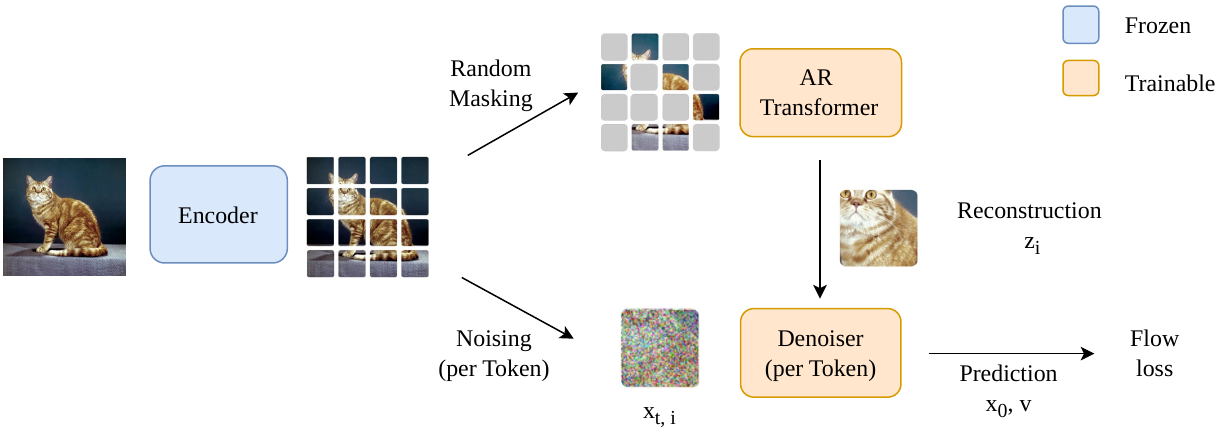}
         \caption{\textbf{Training pipeline.}}
         \vspace{2mm}
         \label{fig:training_pipeline}
     \end{subfigure}
     \begin{subfigure}[t]{.95\linewidth}
         \centering
         \includegraphics[width=.95\linewidth]{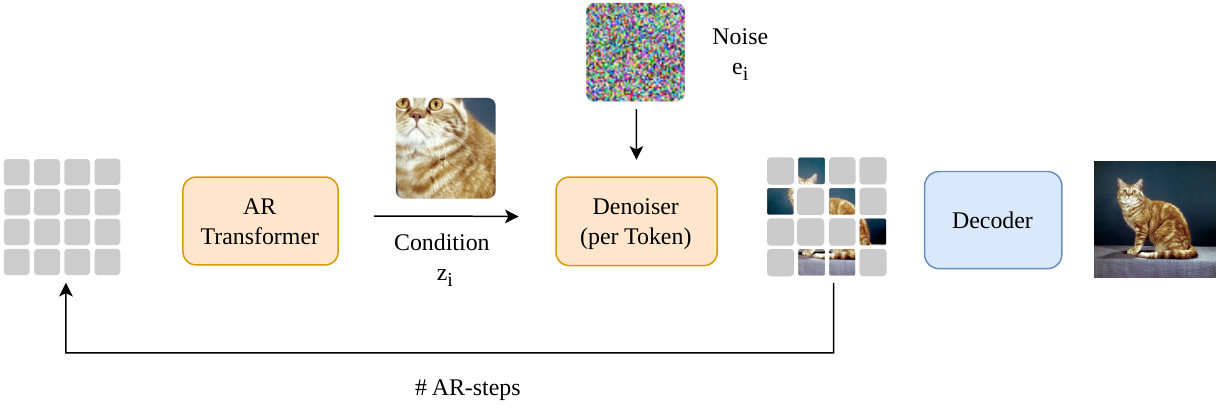}
         \caption{\textbf{Inference pipeline.}}
         \label{fig:generation_pipeline}
     \end{subfigure}
     \caption{
        \textbf{Overview of our masked autoregressive rectified-flow model.} 
        (a) During training, an image is encoded into continuous tokens (or directly patchified in pixel space), randomly masked, and processed by a bidirectional transformer to obtain contextual features $z_i$. A local denoiser predicts the rectified-flow target for each masked token. 
        (b) During sampling, the model starts from a fully masked grid and alternates between context prediction and conditional token generation until all positions are filled. Blue and orange denote frozen and trainable modules, respectively.
     }
\label{fig:pipeline}
\end{figure}

\subsection{Architecture: Context Model and Local Denoiser}
The training and inference pipelines are shown in Figure~\ref{fig:pipeline}. A frozen encoder first maps an image into a token representation. For direct pixel generation, this encoder is replaced by non-overlapping patchification. 

The context transformer follows an MAE-style encoder–decoder structure~\cite{qing2022mar,MaskedAutoencoders2021}. The encoder operates only on visible image tokens, reducing the amount of self-attention computation under high masking. The decoder inserts learned mask tokens and produces the full sequence of contextual vectors $z$. We also evaluate a decoder-only configuration of the context transformer for DINOv2, in which the frozen tokens are supplied directly to the decoder.
Class information is represented by learned in-context tokens and is incorporated by the transformer rather than being supplied directly to the local denoiser. 

The local denoiser is a residual MLP conditioned through adaLN-Zero~\cite{peeblesScalableDiffusionModels2023}. Its input is a single noisy token, while the timestep and transformer context modulate the intermediate activations. Because the denoiser acts independently at each spatial location, it does not contain self-attention; all communication between locations must pass through the transformer-produced context~\cite{qing2022mar}.

We evaluate two modifications of this local denoiser. First, its internal expansion ratio and hidden dimension can be increased to provide more capacity for high-dimensional tokens. Second, the noisy token $x_{t,i}$ and context $z_i$ can be concatenated and projected through an input fusion layer, in addition to using $z_i$ through adaLN modulation. These interventions allow us to distinguish limitations of contextual inference from limitations of the conditional token model.

\section{A Masked-Flow Probe for Target Representations}

\subsection{Target Representations}
We compare four target spaces while fixing the spatial sequence length to 256 tokens. We show t-SNE visualization of these spaces in the appendix.

\noindent\textbf{SD-VAE.}
We use the KL-regularized Stable Diffusion VAE (SD-VAE) with a spatial stride of 16. A 256$\times$256 image is encoded into a 16$\times$16 grid with 16 channels per token.

\noindent\textbf{DINOv2 and MAE}.
We use the base variants of DINOv2 and MAE, both producing 768-dimensional patch features. Their original patch size is 14 at a 224$\times$224 input resolution. Following the setup of RAE~\cite{zhengDiffusionTransformersRepresentation2026}, the spatial features are bilinearly interpolated to obtain a 16$\times$16 token grid for 256$\times$256 images. The global CLS token is discarded, and channel-wise layer normalization is applied to the patch tokens. Frozen pretrained decoders map the resulting features back to pixels. These decoders were trained on ImageNet using a combination of $\ell_1$, LPIPS, and adversarial reconstruction losses~\cite{zhengDiffusionTransformersRepresentation2026}.

\noindent\textbf{Pixels.}
Images are divided into non-overlapping 16$\times$16 RGB patches. Each patch is flattened into a $16 \times 16 \times 3 = 768$ dimensional token. Generated tokens are directly unpatchified into the final image, no pretrained encoder or decoder is used.

\subsection{Autoregressive Sampling}
During inference, generation begins from a fully masked 16$\times$16 token grid, as illustrated in Figure~\ref{fig:pipeline}(b). At each autoregressive step, the transformer processes the currently visible tokens and predicts contextual features for the remaining positions. A subset of masked locations is then activated according to a cosine mask schedule and a random generation order.

For each active location, sampling starts from Gaussian noise and integrates the learned velocity field backward from $t=1$ to $t=0$. We use a deterministic first-order Euler solver. The newly generated tokens are added to the visible context, and the procedure repeats until the complete token grid has been generated. Latent sequences are subsequently decoded to pixels, while pixel tokens are directly unpatchified.

The standard sampling configuration uses 32 autoregressive steps and $50$ flow-integration steps per active token. We additionally evaluate fewer autoregressive and ODE steps, including the extreme case of a single Euler step, to compare how directly different target spaces can be transported from noise to data.

\subsection{Representation-Dependent Guidance}
For class-conditional generation, the model is trained with classifier-free guidance by dropping class-conditioning tokens for $10\%$ of training samples. At inference, we combine the conditional and unconditional velocity estimates as 
\begin{equation}
    \hat{v}_{\Theta} = v_{\Theta}(x_{i, t_i}, t_i, z_{\emptyset}) + w \cdot \big( v_{\Theta}(x_{i, t_i}, t_i, z_i) - v_{\Theta}(x_{i, t_i}, t_i, z_{\emptyset}) \big).
    \label{eq:cfg_guidance_per_token}
\end{equation}
where the conditioning signal $c$ is provided by the context vector $z_i$ and the unconditional null state $z_{\emptyset}$ is represented without class information. $w$ is a guidance scale parameter.

We consider two schedules. \textbf{Linear guidance} applies the same scale (w) throughout generation. \textbf{Progressive guidance} begins close to the unguided prediction and increases the scale as more tokens become visible. This schedule reflects the sequential structure of the generator: early tokens establish the global layout, whereas later tokens refine the image under an increasingly informative context. 
Guidance is treated as an experimental variable rather than a fixed component of the method. We report unguided convergence separately from the best guided generation results, since the evaluated target spaces respond differently to classifier-free guidance.

\section{Representation Choice Changes Optimization}

\subsection{Experimental Setup}
We evaluate generation using Fréchet Inception Distance (FID)~\cite{heusel2017fid}, Inception Score (IS)~\cite{salimans2016improved}, and generative precision and recall. The main model evaluations use 50k generated samples. Guidance sweeps and selected inference ablations are evaluated using 10k samples.
For the pretrained target spaces, we separately evaluate reconstruction quality using LPIPS and reconstruction FID. These metrics characterize the target representation rather than the generative model and are therefore not treated as generation results. 
All models are trained for class-conditional generation on ImageNet~\cite{imagenet} at a resolution of 256$\times$256. The target sequence contains 256 spatial tokens for every representation. More implementation details are provided in the appendix. 

\subsection{Convergence across Target Spaces}
\begin{figure}[t]
     \centering
     \begin{minipage}[c]{0.48\textwidth}
          \centering
          \begin{subfigure}[b]{\textwidth}
              \centering
              \includegraphics[width=\textwidth]{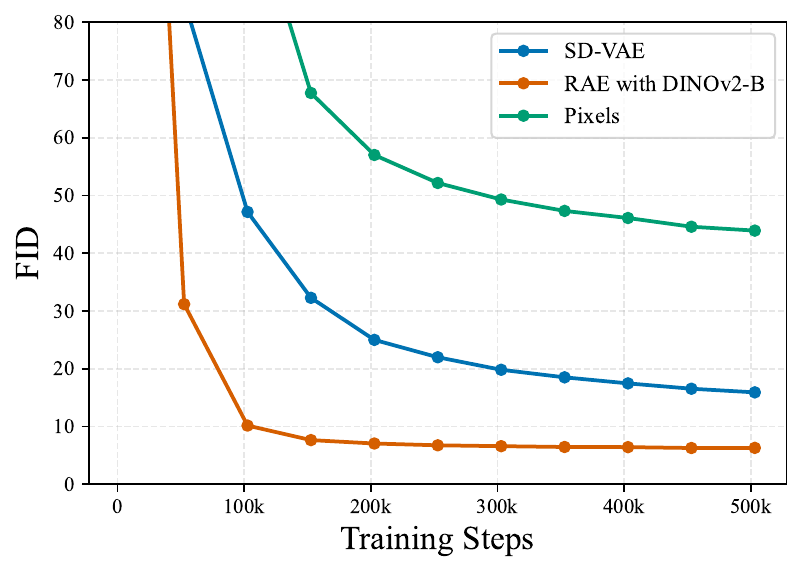}
              \caption{FID over training iterations.}
              \label{fig:comparison_fid_over_steps}
          \end{subfigure}
          \vspace{1.0em}
          \begin{subfigure}[b]{\textwidth}
              \centering
              \includegraphics[width=\textwidth]{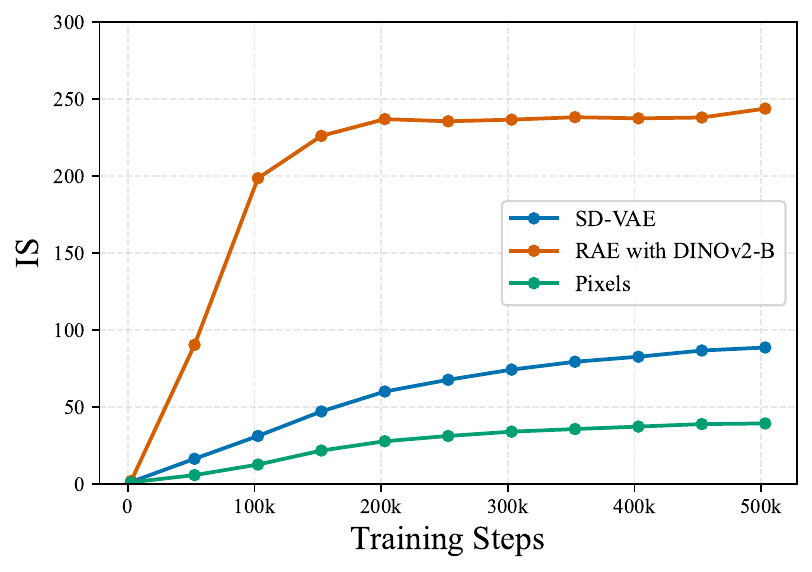}
              \caption{IS over training iterations.}
              \label{fig:comparison_is_over_steps}
          \end{subfigure}
     \end{minipage}
     \hfill
     \begin{minipage}[c]{0.48\textwidth}
          \centering
          \begin{subfigure}[b]{\textwidth}
              \centering
              \includegraphics[width=\textwidth]{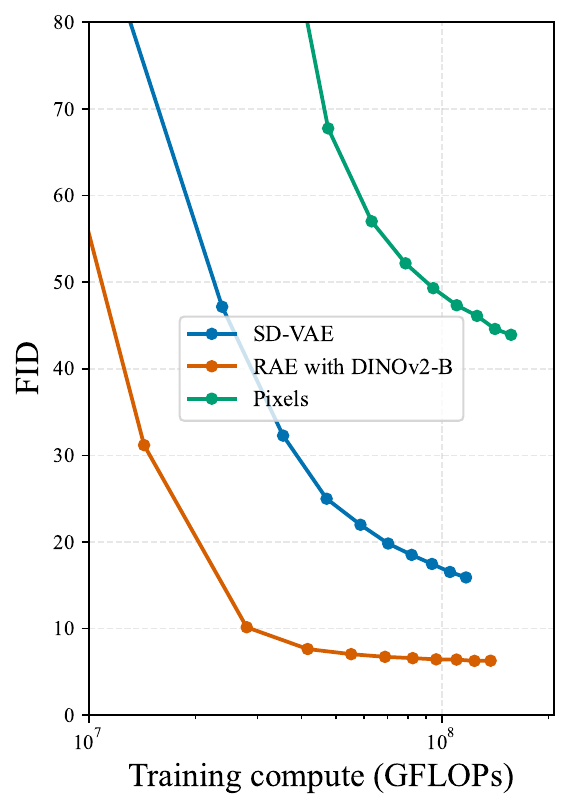}
              \caption{FID vs. accumulated training GFLOPs.}
              \label{fig:gflops}
          \end{subfigure}
     \end{minipage}
     \caption{\textbf{Unguided optimization across target representations.}FID (a) and Inception Score (b) over 503k training steps, and FID versus accumulated training computation (c). DINOv2-B converges fastest under both iteration- and compute-based comparisons; SD-VAE remains intermediate, while the pixel model continues to improve at the end of the shared budget.}
     \label{fig:fid_is}
\end{figure}

We first compare optimization without classifier-free guidance. This avoids conflating representation-dependent training behavior with the large and unequal effects of guidance observed at inference.

Fig.~\ref{fig:fid_is} reports FID and Inception Score as functions of training iterations. DINOv2 improves substantially faster than the compressed SD-VAE and raw pixels. At the end of the shared 503k-step budget, the DINOv2 model reaches an unguided FID of 6.43, compared with 15.98 for SD-VAE and 43.60 for pixels. DINOv2 also obtains the highest unguided Inception Score, 247.63, whereas SD-VAE and pixels obtain 89.53 and 40.19.

The ranking is not explained by parameter count. The pixel model contains 221.6M trainable parameters, more than either latent-space model, but remains substantially slower to optimize. Its FID is still improving at the end of training, indicating that the shared budget is insufficient for convergence.

We additionally plot FID against accumulated training GFLOPs in Fig.~\ref{fig:gflops}. DINOv2 maintains its advantage under this compute-based comparison. SD-VAE remains intermediate, and pixels require substantially more computation to reach the same generation quality. The representation ranking therefore persists when training progress is measured by computation rather than only optimization steps. 

The comparison also shows that target dimensionality alone is not predictive. DINOv2 and pixel tokens are both 768-dimensional, but exhibit the best and worst optimization efficiency, respectively. Conversely, the 16-dimensional SD-VAE target lies between them. 
More details on design choices for adapting the Model to DINOv2 and pixels are provided in the appendix. 

\begin{figure}[!t]
    \centering
    \includegraphics[width=.98\textwidth]{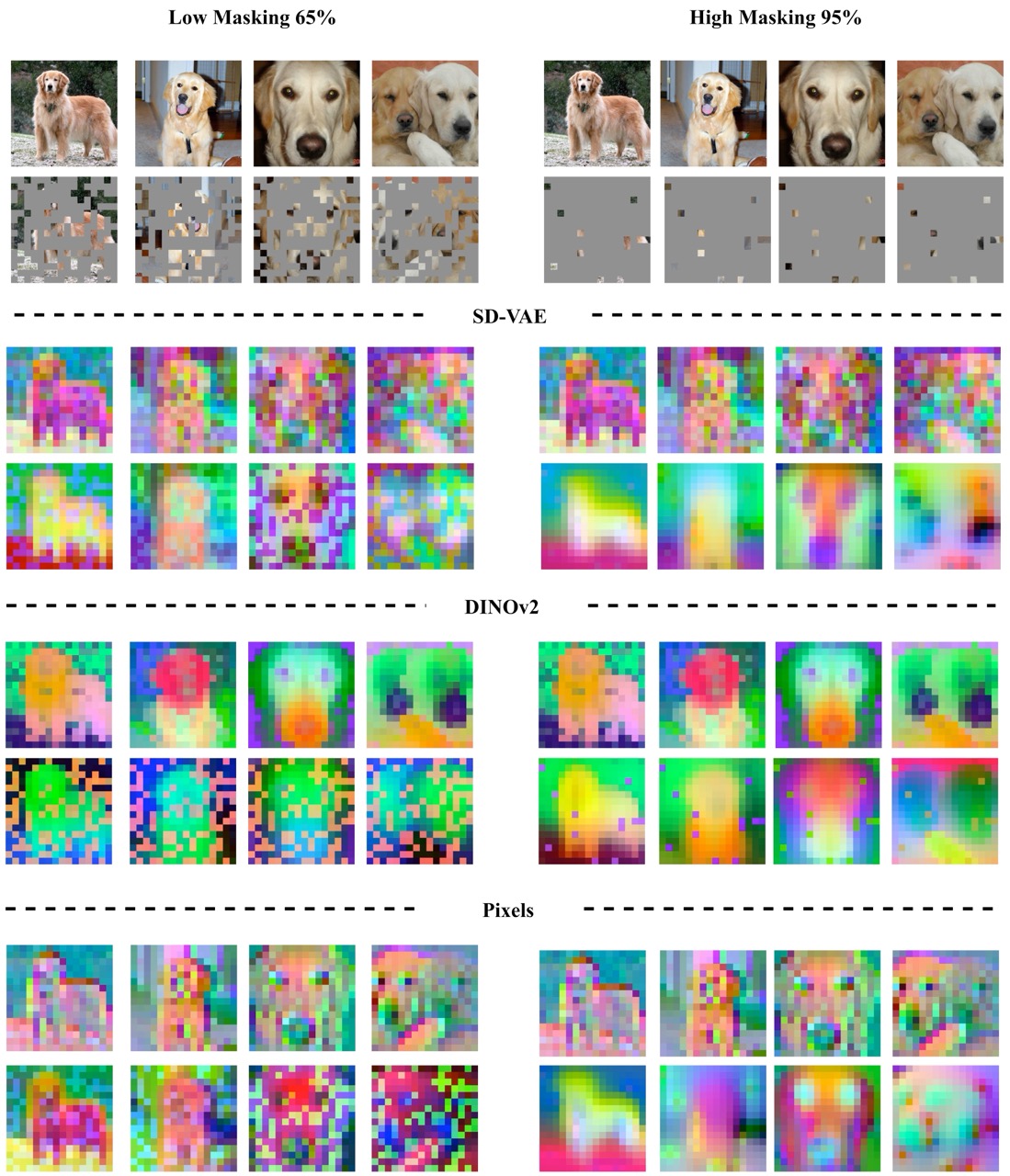} 
    \caption{\textbf{Context reconstruction under 65\% and 95\% masking.} The top rows show the input images and corresponding spatial masks. For each target space, paired rows compare PCA visualizations of the full encoded features and transformer-produced context $z$ under the same mask. At 95\% masking, DINOv2 retains clearer global organization, whereas pixel-space context becomes substantially more degraded. We note that the output of the context transformer on the visible tokens is worse as we do not compute the loss on them, but this does not affect generation as only masked tokens are kept.}
    \label{fig:masking}
\end{figure}
\section{Where Does the Difficulty Move?}

\subsection{Context Reconstruction under Masking}

We examine the contextual features produced by the transformer after 200 epochs of training. We apply the same fixed spatial masks to SD-VAE, DINOv2, and pixel representations at masking ratios of 65\% and 95\%. As shown in Fig.~\ref{fig:masking}, for each representation, we visualize both the fully encoded target features and the transformer output z using PCA coloring.

A common behavior appears across all spaces. The transformer output at visible positions differs from the corresponding original representation, even in the 65\% masking condition. This follows from the training objective: the flow loss is applied only at masked positions, and the transformer is not directly trained to reproduce the visible tokens. These visible-position discrepancies do not affect generation because only the contextual vectors at masked positions condition the local denoiser.

The representations diverge more clearly at 95\% masking. Pixel-space context becomes blurry and loses much of the spatial structure present in the full image. DINOv2 retains a substantially clearer and more semantically coherent contextual organization when only 5\% of the spatial tokens remain visible. SD-VAE lies between these cases.

High masking approximates the early phase of autoregressive sampling, when only a few generated tokens are available. The visualization therefore suggests that DINOv2 provides the transformer with globally informative visible tokens early in generation, whereas localized pixel patches provide less contextual information per observed position.

This interpretation is consistent with the spatial masking behavior of the frozen representations (see appendix Fig.~\ref{fig:comparing_tokenizers}). When individual tokens are masked and the remaining representation is decoded, missing DINOv2 and SD-VAE tokens affect areas extending beyond an isolated patch. MAE masking is more spatially localized, and pixel masking is localized by construction. Equal token masks therefore do not remove equivalent image information across target spaces~\footnote{These visualizations do not provide a direct quantitative measure of conditional entropy or masked-token predictability. They nevertheless expose a representation-dependent difference in the contextual signal available to the local denoiser, particularly in the high-masking regime.}.

\subsection{Distinct Bottlenecks: Context Modeling and Local Denoising}
The DINOv2 architecture reaches its reported performance with a two-layer transformer encoder, whereas the final SD-VAE and pixel configurations use 10-layer encoders. Within DINOv2, a two-layer encoder improves over a decoder-only architecture, but much of the subsequent performance gain comes from the local denoiser rather than additional context-transformer depth. More detailed results are reported in the appendix Tab.~\ref{tab:dino_experiments}. 

Increasing the DINOv2 denoiser expansion ratio and adding direct context fusion improves FID from 7.59 to 6.30. Increasing its hidden dimension to 1536 further improves FID to 5.33. The pixel model shows a related pattern: the bottleneck configuration reaches 71.04, while the wider denoiser and z-fusion reduce FID to 42.39. 

These experiments indicate that high-dimensional target spaces place substantial demands on the local conditional model. A representation may expose useful global structure to the transformer while still requiring a strong denoiser to predict an individual 768-dimensional token from a noisy state and its context.

The experiments do not isolate denoiser width from context fusion because both are changed together in the principal ablations. We therefore attribute the observed improvement to the combined increase in local capacity and conditioning access rather than to either modification alone.

\section{What Does Not Predict Generative Behavior?}
\subsection{Reconstruction Fidelity}

\begin{table}[t]

    \centering
    \small
    \setlength{\tabcolsep}{2pt}
    \renewcommand{\arraystretch}{1.0}

    \caption{
        \textbf{Reconstruction quality of the frozen encoder-decoder pairs on ImageNet validation images. }
        These metrics characterize the target representation and decoder rather than the trained generator. Lower is better for both metrics.
    }

    \label{tab:encoder_reconstruction}

    \begin{tabularx}{\linewidth}{
        X
        >{\small}c
        >{\small}c
        >{\small}c
    }

        \noalign{\hrule height 1pt}
        \noalign{\vskip 0.1em}

        \textbf{Encoder}
        & \textbf{Token dimension}
        & \textbf{LPIPS ($\downarrow$)}
        & \textbf{Reconstruction FID ($\downarrow$)} \\

        \noalign{\hrule height 1pt}
        \noalign{\vskip 0.1em}

        SD-VAE (KL-16)
        & 16
        & 0.14
        & 0.62 \\

        MAE-B
        & 768
        & \textbf{0.11}
        & \textbf{0.16} \\

        \hline

        DINOv2-B
        & 768
        & 0.255
        & 0.49 \\

        DINOv2-L
        & 1024
        & 0.266
        & 0.52 \\

        \noalign{\hrule height 1pt}

    \end{tabularx}

\end{table}

We first evaluate the reconstruction quality of the pretrained target representations. Reconstruction fidelity constrains the information that can appear in the final generated image, but the experiments show that it does not predict how easily the target distribution can be learned.

As shown in Tab.~\ref{tab:encoder_reconstruction}, MAE-B achieves the best reconstruction metrics, with an LPIPS distance of 0.11 and a reconstruction FID of 0.16. SD-VAE follows with an LPIPS of 0.14 and a reconstruction FID of 0.62. DINOv2-B obtains a substantially worse LPIPS distance of 0.255 but a competitive reconstruction FID of 0.49.

Qualitative reconstructions show the same pattern (see appendix Fig.~\ref{fig:reconstructed_images}). MAE preserves fine visual details particularly well, followed by SD-VAE. DINOv2 reconstructions display larger perceptual deviations despite retaining recognizable image content.

However, generation follows a different ranking. DINOv2 reaches substantially better unguided generation than either SD-VAE or MAE, despite not providing the best reconstruction fidelity. MAE provides the strongest reconstructions but yields much worse generation than DINOv2 under the same masked-flow architecture.

Reconstruction fidelity is therefore an important constraint on attainable output quality, but it is not a sufficient predictor of optimization efficiency or final generative performance.

\subsection{DINOv2 vs. MAE: Different Encoder Objectives}
While DINOv2 is trained using a self-supervised objective, MAE utilizes a masking and reconstruction objective. 
Prior work reports that MAE features are less effective than DINOv2 features for joint generative models \cite{zhengDiffusionTransformersRepresentation2026}. We examine whether this ranking persists under masked, per-token flow matching. 

\begin{figure}[t]
     \centering
     \begin{subfigure}[t]{0.48\textwidth}
         \centering
         \includegraphics[width=\textwidth]{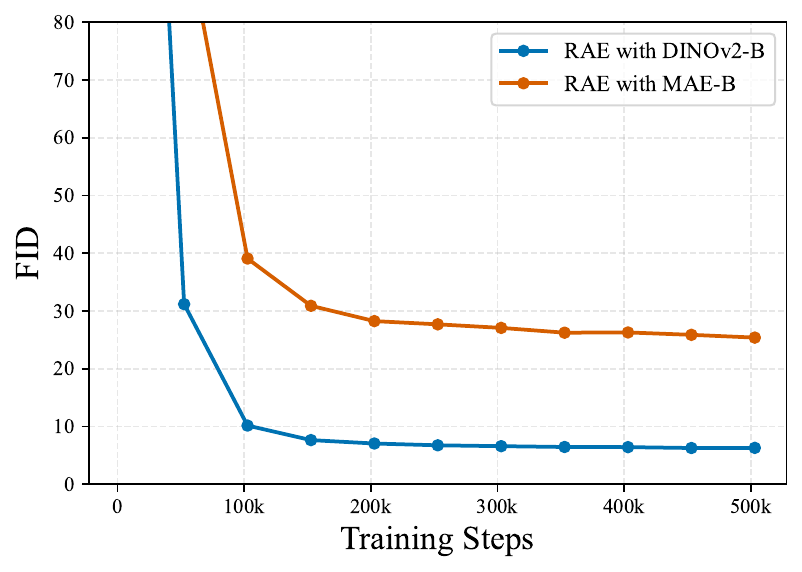}
         \caption{FID over training iterations}
     \end{subfigure}
     \hfill
     \begin{subfigure}[t]{0.48\textwidth}
         \centering
         \includegraphics[width=\textwidth]{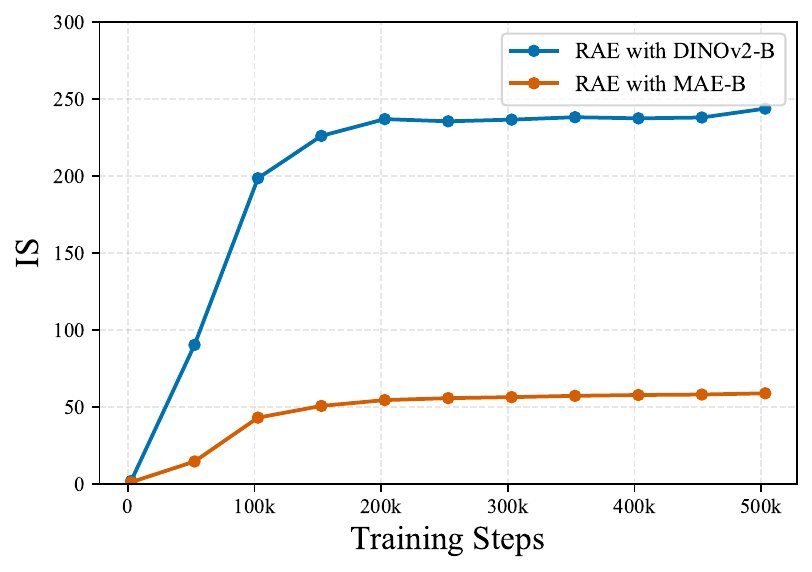}
         \caption{IS over training iterations}
     \end{subfigure}
     \caption{
     \textbf{Generation in DINOv2-B and MAE-B representation spaces. }
     Both models use the same hyperparameters. FID (a) and Inception Score (b) over 200 epochs show substantially better final generation quality in DINOv2 space. }
     \label{fig:dinov2_vs_mae}
\end{figure}

For this experiment, we select the same hyperparameters and architecture as in the final experiment for DINOv2 in Tab.~\ref{tab:cfg}. As shown in Fig.~\ref{fig:dinov2_vs_mae}, we plot the FID and IS metrics over training iterations. The results clearly demonstrate that DINOv2 outperforms MAE features in masked autoregressive models too. We further observe that the convergence speed remains comparable between both spaces, which aligns with expectations given that both latent spaces possess a strong semantic structure.

\section{Representation-Dependent Guidance}
\subsection{Classifier-Free Guidance}

The target spaces respond differently to classifier-free guidance, as shown in Tab.~\ref{tab:cfg}. 
\textbf{SD-VAE} benefits strongly from progressive guidance. The unguided model obtains an FID of 15.98. Increasing the guidance scale at later autoregressive steps improves FID to 3.19 at a maximum scale of $w=6.0$. The associated precision–recall curve exhibits the expected trade-off: stronger guidance increases fidelity and precision while reducing sample coverage.

    

\begin{table}[!tb]

    \centering
    \small
    \setlength{\tabcolsep}{2pt}
    \renewcommand{\arraystretch}{1.0}

    \caption{
        \textbf{Final class-conditional ImageNet generation results after 503k training steps (200 epochs). }
        The upper block reports unguided models; the lower block reports the best observed guidance configuration for each representation. All metrics use 50k generated samples. Symbols (p) and (l) denote progressive and linear classifier-free guidance, respectively.
    }

    \label{tab:cfg}

    \begin{tabularx}{\linewidth}{
        X
        >{\small}c
        >{\small}c
        >{\small}c
        >{\small}c
        >{\small}c
        >{\small}c
    }

        \noalign{\hrule height 1pt}
        \noalign{\vskip 0.1em}

        \textbf{Space}
        & \textbf{Params.}
        & \textbf{CFG scale}
        & \textbf{FID ($\downarrow$)}
        & \textbf{IS ($\uparrow$)}
        & \textbf{Prec. ($\uparrow$)}
        & \textbf{Rec. ($\uparrow$)} \\

        \noalign{\hrule height 1pt}

        \rowcolor{black!10}
        \multicolumn{7}{l}{\textbf{\textit{Without CFG}}} \\

        \quad SD-VAE (KL-16)
        & 179.6M
        & 1.0
        & 15.98
        & 89.53
        & 0.580
        & \textbf{0.606} \\

        \quad DINOv2-B
        & 165.4M
        & 1.0
        & \textbf{6.43}
        & \textbf{247.63}
        & \textbf{0.853}
        & 0.301 \\

        \quad Pixels
        & 221.6M
        & 1.0
        & 43.60
        & 40.19
        & 0.329
        & 0.578 \\

        \noalign{\hrule height 1pt}

        \rowcolor{black!5}
        \multicolumn{7}{l}{\textbf{\textit{With CFG}}} \\

        \quad SD-VAE (KL-16)
        & 179.6M
        & 6.0 (p)
        & \textbf{3.19}
        & \textbf{263.84}
        & 0.757
        & \textbf{0.507} \\

        \quad DINOv2-B
        & 165.4M
        & 1.0 (--)
        & 6.43
        & 247.63
        & \textbf{0.853}
        & 0.301 \\

        \quad Pixels
        & 221.6M
        & 2.0 (l)
        & 8.70
        & 216.76
        & 0.688
        & 0.346 \\

        \noalign{\hrule height 1pt}

    \end{tabularx}

\end{table}

\noindent \textbf{Pixel} generation also depends strongly on guidance. Linear guidance with $w=2.0$ improves the final FID from 43.60 to 8.70. Progressive guidance is less effective, suggesting that the initial pixel patches require strong class conditioning before a reliable global structure has been established.

\noindent \textbf{DINOv2} behaves differently. Neither linear nor progressive guidance improves FID. The best model remains the unguided model with $w=1.0$. This behavior is observed across the evaluated range of guidance scales rather than only at one selected value. 

\noindent \textbf{MAE} depends strongly on guidance, despite using the same representation-autoencoder framework. Its FID improves from 28.00 to 12.69 with linear guidance at $w=2.0$. Progressive guidance does not provide a similar benefit, including at high maximum scales, indicating that strong class conditioning is useful from the early autoregressive steps.

\begin{figure}[t]
     \centering
     \begin{subfigure}[t]{0.48\textwidth}
         \centering
         \includegraphics[width=\textwidth]{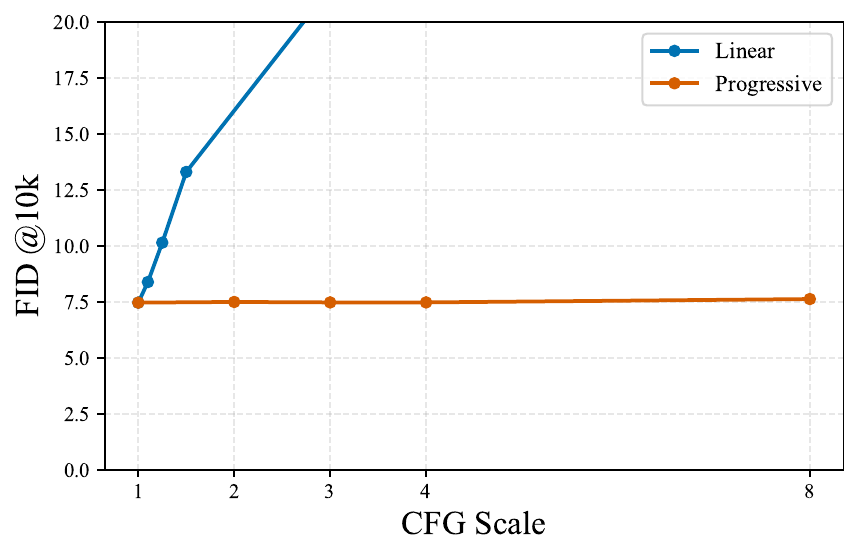}
         \caption{FID @10k, DINOv2 latent space.}
         \label{fig:dinov2_fid_over_cfg}
     \end{subfigure}
     \hfill
     \begin{subfigure}[t]{0.48\textwidth}
         \centering
         \includegraphics[width=\textwidth]{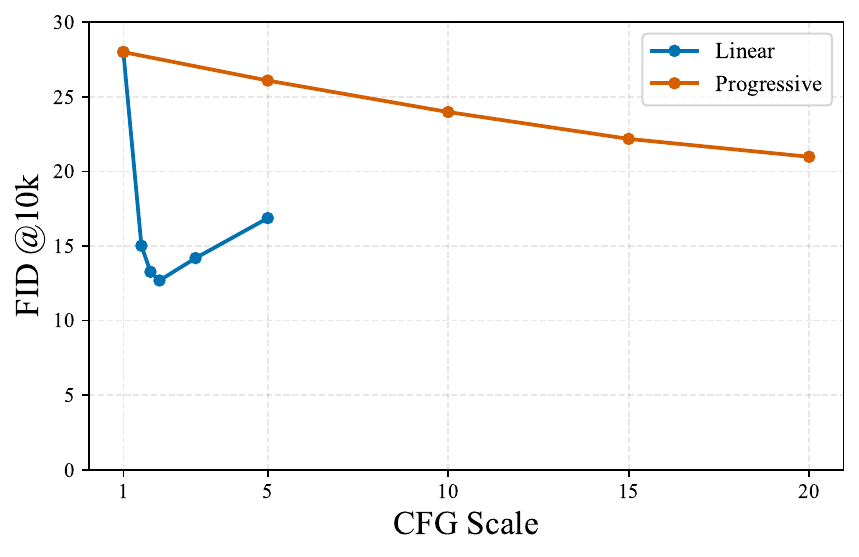}
         \caption{FID @10k, MAE latent space.}
         \label{fig:mae_fid_over_cfg}
     \end{subfigure}
     \caption{\textbf{Representation-dependent response to classifier-free guidance.}
     We report FID@10k under linear and progressive guidance in DINOv2-B (a) and MAE-B (b) representation spaces. DINOv2 obtains its best result without guidance, whereas MAE improves substantially with linear guidance and reaches its best FID at $w=2$.}
     \label{fig:fid_cfg}
\end{figure}

Fig.~\ref{fig:fid_cfg} further shows the complete guidance sweeps for DINOv2 and MAE. DINOv2 degrades under both linear and progressive guidance, whereas MAE benefits from linear guidance applied from the beginning of generation.

These results show that guidance is not a representation-independent inference correction. Even DINOv2 and MAE, which share token dimensionality, spatial layout, and a representation-autoencoder interface, require different guidance strategies.

\subsection{Qualitative Results and Failure Cases}
We show more qualitative samples for all target representations in appendix \ref{app:qualitative_results}. 

\begin{figure}[t]
     \centering
    \begin{subfigure}[t]{0.48\textwidth}
         \centering
         \includegraphics[width=\textwidth]{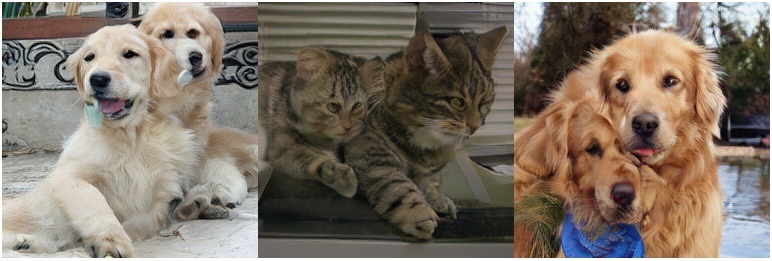}
         \caption{\textbf{DINOv2:} sharp local appearance with failures in global coherence and multi-object separation.}
         \label{fig:failure_cases_dinov2}
     \end{subfigure}
     \hfill
     \begin{subfigure}[t]{0.48\textwidth}
         \centering
         \includegraphics[width=\textwidth]{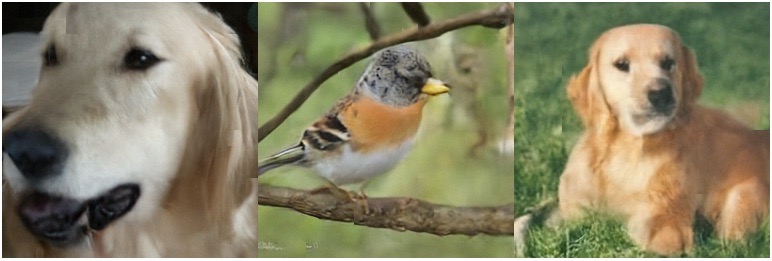}
         \caption{\textbf{Pixels: } recognizable categories and coarse layouts, but limited fine detail.}
         \label{fig:failure_cases_pixels}
     \end{subfigure}
     \caption{\textbf{Representative failure modes of the final DINOv2-B and pixel models.} Representative failure modes of the final DINOv2-B and pixel models.}
     \label{fig:failure_cases}
\end{figure}

DINOv2 generations are often sharp and semantically recognizable even without guidance. What is striking is that the generated images look very natural. However, characteristic failures appear in global structure and object composition. The model can distort object shape or fail to separate multiple semantically related objects. As illustrated in Fig.~\ref{fig:failure_cases_dinov2}, in examples containing two animals, their bodies may merge into one global structure even though local texture and class identity remain plausible.

Pixel-space generations preserve the target class and broad image layout after linear guidance but remain visibly blurrier and contain fewer fine details, as illustrated in Fig.~\ref{fig:failure_cases_pixels}. This limitation persists after clean-data prediction, the 128-dimensional bottleneck, and the wider denoiser. It is consistent with both the incomplete convergence of the pixel model and the remaining burden placed on the local per-token network.

The SD-VAE model produces detailed and diverse samples but requires generation-guidance for high quality results. Here, progressive guidance improves class fidelity and perceptual sharpness while maintaining stronger recall than the other guided models.

\section{Conclusion}
Using a unified masked autoregressive rectified-flow model as a controlled testbed, we compared raw pixels, compressed SD-VAE latents, DINOv2 features, and MAE features as generative targets.
Under a shared ImageNet training budget, the representations produce distinct optimization and inference regimes. 
DINOv2 converges fastest in both iterations and computation, supports recognizable single-step generation, and exposes coherent contextual features under severe masking. Its high-dimensional tokens nevertheless benefit strongly from a wider local denoiser and direct context fusion. Pixels are substantially slower to optimize and require clean-data prediction, a bottleneck, more visible context, a wider denoiser, and linear classifier-free guidance. MAE provides the strongest reconstruction quality and clear semantic organization but remains considerably less generatable than DINOv2. 
These results show that compression, reconstruction fidelity, token dimensionality, and visible semantic clustering do not individually predict masked generative behavior. 
Instead, the target representation changes how modeling difficulty is distributed between contextual inference, conditional token generation, flow integration, and inference-time distributional control.

\clearpage  

\section*{Acknowledgements}
Gege Gao acknowledges the EuroHPC Joint Undertaking for awarding access to computational resources under project ID EHPC-AIF-2026PG01-147 on the Discoverer+ GPU partition hosted by Sofia Tech Park. Andreas Geiger is a member of the Machine Learning Cluster of Excellence, EXC 2064/1, project number 390727645. The authors gratefully acknowledge the ML Cloud and the Tübingen AI Center for providing the computational resources and facilities that supported this work.

%
%
\bibliographystyle{splncs04}
\bibliography{main_cleaned}

\clearpage
\begingroup
\makeatletter
\let\section\tf@llncs@section
\let\subsection\tf@llncs@subsection
\let\subsubsection\tf@llncs@subsubsection
\makeatother
\begin{center}
    {\Large\bfseries Appendix\par}
\end{center}
\vspace{1em}

\appendix

\makeatletter
\renewcommand*{\theHsection}{appendix.\Alph{section}}
\renewcommand*{\theHsubsection}{appendix.\Alph{section}.\arabic{subsection}}
\renewcommand*{\theHsubsubsection}{appendix.\Alph{section}.\arabic{subsection}.\arabic{subsubsection}}
\makeatother

\setcounter{tocdepth}{3}
\makeatletter
\begingroup
\noindent\textbf{\large Table of Contents}\par
\vspace{0.3em}
\hrule
\vspace{0.3em}
\let\oldl@section\l@section
\renewcommand*{\l@section}[2]{\oldl@section{\bfseries #1}{#2}}
\@starttoc{aptoc}
\vspace{0.3em}
\hrule
\endgroup
\let\tf@addcontentsline\addcontentsline
\renewcommand{\addcontentsline}[3]{%
  \tf@addcontentsline{#1}{#2}{#3}%
  \def\tf@tocname{#1}%
  \def\tf@targettoc{toc}%
  \ifx\tf@tocname\tf@targettoc
    \addtocontents{aptoc}{\protect\contentsline{#2}{#3}{\thepage}{\@currentHref}}%
  \fi
}
\makeatother

\newpage

\section{The Storyline}

\begingroup
\footnotesize

\setlist[enumerate,1]{
    label=\arabic*.,
    leftmargin=1.6em,
    itemsep=0.35em,
    topsep=0.2em,
    parsep=0pt
}

\setlist[enumerate,2]{
    label=\alph*.,
    leftmargin=1.5em,
    itemsep=0.12em,
    topsep=0.08em,
    parsep=0pt
}

\begin{enumerate}

\item \textbf{Why interesting?}

\begin{enumerate}

    \item Training modern image generators requires substantial
    \firstterm{training computation};
    the \firstterm{target representation}
    directly defines the distribution that the generator must learn.

    \item In \firstterm{masked generative models},
    each prediction combines two distinct problems:
    inferring context from partially observed tokens,
    which we call \firstterm{contextual inference},
    and modeling the conditional distribution of a missing token,
    which we call \firstterm{conditional token modeling}.

    \item Understanding how the
    \reuseterm{target representation}
    affects these two problems can guide the joint choice of
    \firstterm{model capacity},
    \firstterm{masking strategy}, and
    \firstterm{sampling procedure},
    rather than relying on representation-specific trial and error.

\end{enumerate}

\item \textbf{How done now?}

\begin{enumerate}

    \item Most diffusion and continuous
    \reuseterm{masked generative models}
    operate in compressed, reconstruction-trained
    \firstterm{SD-VAE latent spaces}.

    \item Recent \firstterm{joint diffusion and flow models}
    instead generate
    \firstterm{semantic representation-autoencoder features}
    or \firstterm{raw pixels},
    and report improved optimization under suitable architectures.

    \item Continuous
    \reuseterm{masked generative models}
    use a \firstterm{transformer}
    to perform \reuseterm{contextual inference}
    and a \firstterm{local diffusion or flow head}
    to perform \reuseterm{conditional token modeling},
    but alternative
    \reuseterm{target representations}
    remain poorly understood in this conditional setting.

\end{enumerate}

\item \textbf{What is missing, and So What?}

\begin{enumerate}

    \item Results from
    \reuseterm{joint generation}
    do not directly determine which
    \reuseterm{target representations}
    are easy to model under
    \reuseterm{partial observation}.

    \item Common explanations---
    \firstterm{compression},
    \firstterm{reconstruction fidelity},
    \firstterm{high-level semantics}, or
    \firstterm{low-dimensional data structure}---
    remain entangled and are individually insufficient to predict
    \firstterm{generative behavior}.

    \item Treating
    \reuseterm{target representations}
    as interchangeable obscures where
    \reuseterm{model capacity}
    is required:
    in \reuseterm{contextual inference},
    in \reuseterm{conditional token modeling},
    or during
    \firstterm{inference-time distributional control}.

    \item Consequently, architectures,
    \reuseterm{masking strategies}, and
    \firstterm{guidance strategies}
    optimized for one
    \reuseterm{target representation}
    may transfer poorly to another;
    faster convergence may also conceal reduced
    \firstterm{distribution coverage}.

\end{enumerate}

\item \textbf{Proposed approach.}

\begin{enumerate}

    \item Use one
    \firstterm{masked autoregressive generator}
    with a
    \firstterm{per-token rectified-flow objective}
    as a
    \firstterm{common experimental probe}.

    \item Compare
    \reuseterm{raw pixels},
    \reuseterm{SD-VAE latent spaces},
    \firstterm{DINOv2 representation-autoencoder features}, and
    \firstterm{MAE representation-autoencoder features}
    under a shared dataset,
    \firstterm{training budget},
    \firstterm{sequence length}, and
    \firstterm{model scale}.

    \item Decompose generative difficulty through interventions on
    \reuseterm{contextual inference},
    \firstterm{local denoiser capacity},
    \firstterm{flow integration},
    \reuseterm{masking strategy}, and
    \firstterm{classifier-free guidance}.

    \item Test the empirical hypothesis that
    \reuseterm{target representations}
    reallocate, rather than uniformly remove,
    difficulty across these components.

\end{enumerate}

\item \textbf{Experimental questions.}

\begin{enumerate}[label={}, leftmargin=0pt, itemsep=0.25em]

    \item
    \textbf{C1. Is the
    \reuseterm{common experimental probe}
    reasonable?}
    Can the same
    \firstterm{masked-flow formulation}
    train viable generators in
    \reuseterm{compressed latent spaces},
    \reuseterm{semantic representation-autoencoder features}, and
    \reuseterm{raw pixels}?

    \item
    \textbf{C2. How does the
    \reuseterm{target representation}
    affect optimization?}
    Under matched
    \reuseterm{training budgets}
    and similar
    \reuseterm{model scales},
    which \reuseterm{target representations}
    converge fastest in iterations and total
    \reuseterm{training computation}?

    \item
    \textbf{C3. Where is
    \reuseterm{contextual inference}
    difficult?}
    How coherent is the inferred context under severe
    \reuseterm{masking},
    and how much
    \reuseterm{model capacity}
    must be allocated to the
    \reuseterm{transformer}
    in each space?

    \item
    \textbf{C4. Where is
    \reuseterm{conditional token modeling}
    difficult?}
    How do
    \reuseterm{local denoiser capacity},
    \firstterm{conditioning},
    \firstterm{prediction parameterization}, and
    \firstterm{the number of ODE steps}
    affect different
    \reuseterm{target representations}?

    \item
    \textbf{C5. Are common representation properties
    sufficient explanations?}
    Do
    \reuseterm{compression},
    \reuseterm{reconstruction fidelity}, or
    \firstterm{visible semantic clustering}
    consistently predict
    \firstterm{generative learnability},
    particularly when comparing
    \reuseterm{DINOv2 representation-autoencoder features}
    and
    \reuseterm{MAE representation-autoencoder features}?

    \item
    \textbf{U1. How does the
    \reuseterm{target representation}
    reshape the
    \firstterm{generated distribution}?}
    How do
    \reuseterm{target representations}
    change
    \reuseterm{classifier-free guidance},
    \firstterm{precision--recall},
    \firstterm{sample diversity}, and
    \firstterm{characteristic failure modes}?

\end{enumerate}

\end{enumerate}

\section{Related Work}
\subsection{Generative Models across Target Representations}
Diffusion and flow-matching models learn a continuous trajectory between a simple noise distribution and the data distribution. Early diffusion models formulate this trajectory as a stochastic forward and reverse process \cite{sohl-dicksteinDeepUnsupervisedLearning2015,hoDenoisingDiffusionProbabilistic2020}, while deterministic formulations such as DDIM and probability-flow ODEs enable sampling with fewer steps \cite{songDenoisingDiffusionImplicit2021,songScoreBasedGenerativeModeling2021}. Flow matching instead directly regresses the velocity field of a prescribed probability path \cite{lipmanFlowMatchingGenerative2023,liuFlowStraightFast2023}. Rectified flow adopts a linear interpolation between data and Gaussian noise, providing a simple training objective and deterministic ODE sampling \cite{liuFlowStraightFast2023,esserScalingRectifiedFlow2024}.

The target distribution learned by these models is commonly defined through a pretrained autoencoder. Latent diffusion models use a reconstruction-trained variational autoencoder to remove perceptually less important image information and reduce both spatial resolution and channel dimensionality \cite{rombachHighResolutionImageSynthesis2022}. Stable Diffusion VAE representations have subsequently become a standard target for diffusion transformers and continuous autoregressive image generators \cite{peeblesScalableDiffusionModels2023,maSiTExploringFlow2024,liAutoregressiveImageGeneration2024}.

Recent work has investigated alternative representations. Representation autoencoders pair a frozen vision foundation model with a separately trained image decoder \cite{zhengDiffusionTransformersRepresentation2026,kocabas2026trajectoryforcing}. Compared with conventional VAEs, these encoders are not trained primarily to reconstruct pixels. Instead, they inherit objectives such as self-distillation, masked reconstruction, or image-text alignment. Joint diffusion and flow models trained on such representations can converge rapidly and achieve strong generation quality, although their reconstruction properties differ substantially across encoders \cite{zhengDiffusionTransformersRepresentation2026,singhImprovedBaselinesRepresentation2026,bflRepresentationComparison2025}.

A related line of work aligns intermediate generative features with pretrained visual representations rather than directly generating those representations. REPA aligns diffusion-transformer features with DINOv2 features \cite{yuRepresentationAlignmentGeneration2025}, while Self-Flow combines representation learning and generation through a self-supervised teacher-student objective \cite{cheferSelfSupervisedFlowMatching2026}. These approaches show that semantic visual representations can improve generative optimization, but they do not determine how such representations behave when used directly as the output space of a masked conditional generator.

At the opposite extreme, recent models revisit direct pixel generation. JiT shows that a denoising model can operate in high-dimensional pixel space when it predicts clean data and uses a capacity-limited bottleneck \cite{liBackBasicsLet2026}. Other work similarly explores latent-free or pixel-space diffusion and flow models \cite{yuPixelDiTPixelDiffusion2025,luOnestepLatentfreeImage2026,baadeLatentForcingReordering2026}. Direct pixels avoid the reconstruction ceiling of a pretrained decoder, but present a high-dimensional and spatially local target distribution.

We study these target spaces within one masked autoregressive flow formulation. In contrast to joint diffusion or flow models, each prediction is conditioned on a partially observed token sequence. The target representation therefore affects not only the marginal distribution to be generated, but also the information available through visible tokens and the conditional distribution of each missing token.

\subsection{Continuous Masked Image Generation}
Autoregressive image models decompose image generation into conditional predictions. Early pixel autoregressive models use a fixed spatial order \cite{oordConditionalImageGeneration2016,chenGenerativePretrainingPixels2020}, while later approaches generate discrete visual tokens produced by vector-quantized autoencoders \cite{esserTamingTransformersHighResolution2021,rameshZeroShotGeneration2021}. Such models benefit from the scalability of next-token prediction, but a fixed causal order limits parallel generation and does not naturally exploit two-dimensional context.

Masked generative models instead use bidirectional attention and predict randomly hidden tokens from the visible image context. MaskGIT demonstrates that multiple discrete tokens can be generated in parallel according to an iterative masking schedule \cite{changMaskGITMaskedGenerative2022}. MAGE further combines masked representation learning and generation within one architecture \cite{liMAGEMAskedGenerative2023}.

MAR extends masked autoregression to continuous-valued tokens \cite{liAutoregressiveImageGeneration2024}. Rather than applying a categorical output head, MAR uses a small diffusion network to model the conditional distribution of each missing token. The transformer models dependencies across token positions, while a local denoiser models the continuous distribution $p(x_i \mid z_i)$. This separation makes it possible to generate continuous autoencoder features without vector quantization.

Subsequent work improves the efficiency of continuous masked generation through alternative masking schedules, architectures, and loss formulations \cite{youEffectiveEfficientMasked2025}. However, these models are predominantly developed and evaluated in SD-VAE latent space. We retain their separation between global contextual modeling and local token generation, but replace the local diffusion objective with rectified flow and systematically vary the target representation.

\section{Experimental Protocol}

\subsection{Dataset and Evaluation}
All models are trained for class-conditional generation on ImageNet~\cite{imagenet} at a resolution of 256$\times$256. The target sequence contains 256 spatial tokens for every representation. 

We evaluate generation using Fréchet Inception Distance (FID)~\cite{heusel2017fid}, Inception Score (IS)~\cite{salimans2016improved}, and generative precision and recall. The main model evaluations use 50k generated samples. Guidance sweeps and selected inference ablations are evaluated using 10k samples.
For the pretrained target spaces, we separately evaluate reconstruction quality using LPIPS and reconstruction FID. These metrics characterize the encoder-decoder pair rather than the generative model and are therefore not treated as generation results. 

\subsection{Shared Training Budget}
The transformer hidden dimension is 768 with 12 attention heads, and the local denoiser contains six residual MLP blocks with a hidden dimension of 1024 in the principal cross-representation comparison. The final SD-VAE, DINOv2-B, and pixel configurations contain 179.6M, 165.4M, and 221.6M trainable parameters, respectively. The models are therefore approximately, but not exactly, matched in scale. The DINOv2 analysis additionally includes a 255.1M-parameter model with a 1536-dimensional denoiser. This configuration is used only to study the effect of local denoiser capacity. The cross-representation efficiency comparison instead uses the 165.4M-parameter DINOv2 model to remain closer to the scale of the SD-VAE baseline. 

All final models are trained for 200 epochs, corresponding to approximately 500k optimization steps, with a batch size of 512. We use Adam optimizer with $(\beta_1,\beta_2)=(0.9,0.95)$, a constant learning rate of $2\times10^{-4}$, weight decay of 0.02, 50 warm-up epochs, and an exponential-moving-average decay of 0.9999.

\subsection{Representation-Specific Configurations}
The shared architecture provides a common generative factorization, but the final training configuration is not identical across all target spaces. We evaluate representation-specific timestep distributions, masking distributions, prediction parameterizations, and denoiser capacities.

\noindent\textbf{SD-VAE model} uses velocity prediction, logit-normal timestep sampling, and an exponential mask-rate distribution. Its transformer contains 10 encoder and 10 decoder layers, and its denoiser uses an expansion ratio of one.

\noindent\textbf{DINOv2-B model} uses velocity prediction, a shifted logit-normal timestep distribution, and the same exponential mask-rate family. Its contextual encoder is reduced to two layers while retaining a 10-layer decoder. The local denoiser uses a larger expansion ratio of four and direct fusion between the noisy token and contextual feature.

\noindent\textbf{Pixel model} uses clean-data prediction, a 128-dimensional bottleneck at the denoiser input, and a less aggressive truncated-normal mask-rate distribution with a lower bound of 0.7. Its transformer contains 10 encoder and 10 decoder layers, and its denoiser uses an expansion ratio of four with contextual input fusion.

\noindent\textbf{DINOv2-MAE comparison} uses the same generative architecture and hyperparameter configuration for both feature encoders. It therefore isolates the encoder representation more closely than comparisons involving SD-VAE or pixels, although the pretrained encoder-decoder systems themselves remain different.

These representation-specific choices are part of the empirical findings: \textbf{a single schedule and architecture did not transfer equally well across target spaces.} At the same time, they limit a strictly causal interpretation of the final cross-space rankings. We therefore use two complementary forms of evidence. Unguided convergence under a common training budget measures the efficiency of the resulting representation-specific systems, while within-space interventions identify which components are responsible for substantial performance changes.

\begin{table}[t]
    \centering
    \scriptsize
    \setlength{\tabcolsep}{2pt}
    \renewcommand{\arraystretch}{1.0}
    
    \caption{Comparison of the target representation spaces.}
    \label{tab:target_spaces}
    \begin{tabularx}{\linewidth}{
        >{\raggedright\arraybackslash}p{2.4cm}
        >{\centering\arraybackslash}p{1.2cm}
        >{\centering\arraybackslash}p{1.2cm}
        >{\raggedright\arraybackslash}X
        >{\raggedright\arraybackslash}p{3.0cm}
    }
        \toprule
        \textbf{Target space}
        & \textbf{Grid}
        & \textbf{Dim.}
        & \textbf{Encoder obj.}
        & \textbf{Pixel reconstruction} \\
        \midrule

        SD-VAE (KL-16)
        & $16 \times 16$
        & 16
        & Recon. loss + KL loss
        & Frozen SD-VAE decoder \\

        DINOv2-B
        & $16 \times 16$
        & 768
        & Self-supervision + distillation
        & Frozen RAE decoder \\

        MAE-B
        & $16 \times 16$
        & 768
        & Masked patch recon.
        & Frozen RAE decoder \\

        Pixels
        & $16 \times 16$
        & 768
        & None
        & Direct unpatchification \\

        \bottomrule
    \end{tabularx}
\end{table}

\subsubsection{(i) Adapting the Model to DINOv2. }

We identify which design choices improve generation in DINOv2 space. As shown in Tab.~\ref{tab:dino_experiments}, all models in this ablation are trained for 503k steps and evaluated using 50k generated samples without classifier-free guidance.

\begin{table}[!tbh]

    \centering
    \small
    \setlength{\tabcolsep}{2pt}
    \renewcommand{\arraystretch}{1.0}

    \caption{
        \textbf{Step-by-step comparison of design choices in the DINOv2 latent space.}
        Metrics are reported on 50k generated samples.
    }

    \label{tab:dino_experiments}

    \begin{tabularx}{\linewidth}{
        X
        >{\small}c
        >{\small}c
        >{\small}c
        >{\small}c
    }

        \noalign{\hrule height 1pt}
        \noalign{\vskip 0.1em}

        \textbf{Space: DINOv2}
        & \textbf{Params}
        & \textbf{Epochs}
        & \textbf{FID ($\downarrow$)}
        & \textbf{IS ($\uparrow$)} \\

        \noalign{\hrule height 1pt}

        \rowcolor{black!5}\multicolumn{5}{l}{\textbf{\textit{Prediction type}}} \\

        \quad v-prediction
        & 125.6M
        & 200
        & 9.54
        & 186.85 \\

        \quad x-prediction
        & 125.6M
        & 200
        & 10.70
        & 158.70 \\

        \noalign{\hrule height 1pt}

        \rowcolor{black!5}\multicolumn{5}{l}{\textbf{\textit{Timestep schedule}}} \\

        \quad uniform time sampling
        & 125.6M
        & 200
        & 9.54
        & 186.85 \\

        \quad logit-normal + shift ($\alpha=6.93$)
        & 125.6M
        & 200
        & 7.59
        & 211.37 \\

        \noalign{\hrule height 1pt}

        \rowcolor{black!5}\multicolumn{5}{l}{\textbf{\textit{Architecture}}} \\

        \quad decoder-only
        & 124.7M
        & 200
        & 8.55
        & 204.48 \\

        \quad encoder (2 layer) + decoder
        & 125.6M
        & 200
        & 7.59
        & 211.37 \\

        \noalign{\hrule height 1pt}

        \rowcolor{black!5}\multicolumn{5}{l}{\textbf{\textit{Denoiser conditioning}}} \\

        \quad adaLN (dim=1024, ratio 1)
        & 124.6M
        & 200
        & 7.59
        & 211.37 \\

        \quad adaLN (dim=1024, ratio $4$) + z-fusion
        & 165.4M
        & 200
        & 6.30
        & 243.91 \\

        \quad adaLN (dim=1536, ratio $4$) + z-fusion
        & 255.1M
        & 200
        & 5.33
        & 248.00 \\

        \noalign{\hrule height 1pt}

    \end{tabularx}

\end{table}

We begin with a 125.6M-parameter model using two transformer-encoder layers and a local denoiser with hidden dimension 1024 and expansion ratio one. With uniform timestep sampling, velocity prediction reaches an FID of 9.54 and an Inception Score of 186.85. Clean-data prediction does not improve the result, obtaining an FID of 10.70 and an Inception Score of 158.70. We therefore retain velocity prediction for DINOv2.

Replacing uniform timestep sampling with a shifted logit-normal distribution improves FID from 9.54 to 7.59 and Inception Score from 186.85 to 211.37. The shift parameter is $\alpha=6.93$, which allocates greater training density toward the high-noise part of the trajectory.

We also compare a decoder-only context transformer with a shallow encoder-decoder model. The decoder-only model reaches an FID of 8.55, while adding a two-layer encoder improves FID to 7.59. The shallow encoder therefore remains useful even though the frozen DINOv2 tokens already contain semantic information. Because the encoder also processes the learned class tokens jointly with the visible image features, the result may reflect both additional contextual processing and improved class-conditioning integration.

The largest improvement comes from modifying the local token model. Increasing the denoiser expansion ratio from one to four and adding direct z-fusion improves FID from 7.59 to 6.30 and Inception Score from 211.37 to 243.91. Increasing the denoiser hidden dimension further, from 1024 to 1536, reaches an FID of 5.33 and an Inception Score of 248.00.

These results show that the rapid optimization of DINOv2 does not eliminate the need for a capable conditional token model. Its context can be processed with a relatively shallow encoder, but the 768-dimensional targets benefit strongly from a wider local denoiser and direct context-token fusion.

For the principal comparison with SD-VAE and pixels, we use the 165.4M-parameter configuration rather than the larger 255.1M model.

\subsubsection{(ii) Adapting the Model to Pixels. }

Pixel-space generation requires a different sequence of adaptations. We begin with the SD-VAE-style architecture and train all configurations for 503k steps without classifier-free guidance.

\begin{table}[t]

    \centering
    \small
    \setlength{\tabcolsep}{2pt}
    \renewcommand{\arraystretch}{1.0}

    \caption{
        \textbf{Step-by-step comparison of design choices in pixel space.}
        Metrics are reported on 50k generated samples.
    }

    \label{tab:pixel_experiments}

    \begin{tabularx}{\linewidth}{
        X
        >{\small}c
        >{\small}c
        >{\small}c
        >{\small}c
    }

        \noalign{\hrule height 1pt}
        \noalign{\vskip 0.1em}

        \textbf{Space: Pixel}
        & \textbf{Params}
        & \textbf{Epochs}
        & \textbf{FID ($\downarrow$)}
        & \textbf{IS ($\uparrow$)} \\

        \noalign{\hrule height 1pt}

        \rowcolor{black!5}
        \multicolumn{5}{l}{\textbf{\textit{Prediction type}}} \\

        \quad v-prediction
        & 182.3M
        & 200
        & 104.69
        & 15.82 \\

        \quad x-prediction
        & 182.3M
        & 200
        & 89.64
        & 21.10 \\

        \noalign{\hrule height 1pt}

        \rowcolor{black!5}
        \multicolumn{5}{l}{\textbf{\textit{Mask-rate schedule}}} \\

        \quad exponential
        & 182.3M
        & 200
        & 89.64
        & 21.10 \\

        \quad truncated normal
        & 182.3M
        & 200
        & 85.05
        & 23.36 \\

        \noalign{\hrule height 1pt}

        \rowcolor{black!5}
        \multicolumn{5}{l}{\textbf{\textit{Bottleneck layer}}} \\

        \quad no bottleneck
        & 182.3M
        & 200
        & 85.05
        & 23.36 \\

        \quad bottleneck $d'=128$
        & 182.3M
        & 200
        & 71.04
        & 24.62 \\

        \noalign{\hrule height 1pt}

        \rowcolor{black!5}
        \multicolumn{5}{l}{\textbf{\textit{Denoiser conditioning}}} \\

        \quad adaLN (dim=1024, ratio 1)
        & 182.3M
        & 200
        & 71.04
        & 24.62 \\

        \quad adaLN (dim=1024, ratio $4$) + z-fusion
        & 221.6M
        & 200
        & 42.39
        & 41.04 \\

        \noalign{\hrule height 1pt}

    \end{tabularx}

\end{table}

As reported in Tab.~\ref{tab:pixel_experiments}, with velocity prediction, the initial pixel model obtains an FID of 104.69 and an Inception Score of 15.82. Unlike the result in DINOv2 space, clean-data prediction is beneficial for pixels: it improves FID to 89.64 and Inception Score to 21.10.

We next replace the aggressive exponential mask-rate distribution with a truncated-normal distribution. This gives the model more visible context on average and improves FID from 89.64 to 85.05. The improvement is modest but consistent with the highly localized nature of pixel patches: an individual patch contains no feature information from neighboring regions, unlike tokens produced by a vision transformer or convolutional autoencoder.

Introducing a 128-dimensional bottleneck at the denoiser input yields a larger improvement, reducing FID from 85.05 to 71.04. The bottleneck maps each 768-dimensional pixel patch into a lower-dimensional intermediate space before expanding it into the denoiser hidden dimension.

Finally, increasing the denoiser expansion ratio from one to four and adding direct z-fusion improves FID from 71.04 to 42.39 and Inception Score from 24.62 to 41.04. As in DINOv2 space, the high-dimensional target benefits substantially from a stronger local model and direct access to transformer context.

The pixel ablation demonstrates that the target space does not only change convergence speed. It changes which modeling choices are effective. Clean-data prediction and a bottleneck improve pixel generation, whereas clean-data prediction does not help DINOv2. Pixels benefit from a less aggressive mask distribution, while DINOv2 uses the exponential masking family. Both high-dimensional spaces benefit from a stronger denoiser and direct contextual fusion.

\subsection{Detailed Parameters and Architectures}
In Tab.~\ref{tab:details}, we present the architectural, training and sampling configurations used in our experiments for all representation spaces. The Table is structurally categorized into neural network architectures, training optimization details, and specific inference sampling details.

\begin{table}[t]
    \centering
    \caption{Implementation details for all representation spaces. }
    \label{tab:details}
    
    \resizebox{\textwidth}{!}{
    \begin{tabular}{l|c|c|c}
        \toprule
        & \multicolumn{3}{c}{\textbf{FlowMAR-B}} \\
        \cmidrule(lr){2-4}
        \textbf{Repr. Space} & \textbf{SD-VAE} & \textbf{DINOv2-B} & \textbf{Pixels} \\
        \hline
        \rowcolor{gray!20} \textbf{architecture} & & & \\
        parameters (M) & 179.6M & 165.4M & 221.6M \\
        AR-trans. encoder depth & 10 & 2 & 10 \\
        AR-trans. decoder depth & 10 & 10 & 10 \\
        AR-trans. hidden dim & 768 & 768 & 768 \\
        AR-trans. heads & 12 & 12 & 12 \\
        AR-trans. architecture & MAE-style & MAE-style & MAE-style \\
        AR-trans. dropout & 0.1 & 0.1 & 0.1 \\
        AR-trans. class tokens & 64 & 64 & 64 \\
        denoiser depth & 6 & 6 & 6 \\
        denoiser hidden dim & 1024 & 1024 & 1024 \\
        denoiser block ratio & 1 & 4 & 4 \\
        denoiser dropout & 0.0 & 0.0 & 0.0 \\
        image size & 256 & 256 & 256 \\
        patch size & 1 & 1 & image / 16 \\
        bottleneck & - & - & 128 \\
        \hline
        \rowcolor{gray!20} \textbf{training} & & & \\
        epochs & 200 & 200 & 200 \\
        warmup epochs & 50 & 50 & 50 \\
        optimizer & Adam, $\beta_1, \beta_2 = 0.9, 0.95$ & Adam, $\beta_1, \beta_2 = 0.9, 0.95$ & Adam, $\beta_1, \beta_2 = 0.9, 0.95$ \\
        batch size & 512 & 512 & 512 \\
        learning rate & 2e-4 & 2e-4 & 2e-4 \\
        learning rate schedule & constant & constant & constant \\
        weight decay & 0.02 & 0.02 & 0.02 \\
        ema decay & 0.9999 & 0.9999 & 0.9999 \\
        time sampler & logit-normal & logit-normal + shift & logit-normal \\
        mask rate sampler & Exp + clamb 0.62 & Exp + clamb 0.62 & trunc normal 0.7 \\
        class token drop (for CFG) & 0.1 & 0.1 & 0.1 \\
        \hline
        \rowcolor{gray!20} \textbf{sampling} & & & \\
        ODE solver & Euler & Euler & Euler \\
        ODE steps & 50 & 50 & 50 \\
        time steps & linear in [0.0, 1.0] & shifted in [0.0, 1.0] & linear in [0.0, 1.0] \\
        mask schedule & cosine & cosine & cosine \\
        AR steps & 32 & 32 & 32 \\
        generation order & random & random & random \\
        \bottomrule
    \end{tabular}
    }
\end{table}

\section{More on ``Where Does the Difficulty Move?''}
\subsection{Autoregressive Sampling Steps}
We vary the number of autoregressive steps while keeping the trained models fixed. Because every image contains 256 tokens, fewer autoregressive steps require more tokens to be generated in parallel.

As shown in Fig.~\ref{fig:comparison_ar_steps}, across SD-VAE, DINOv2, and pixels, generation quality approaches convergence at approximately 32 autoregressive steps. At this setting, the model generates an average of eight new tokens per step. DINOv2 and SD-VAE also retain acceptable qualitative results with only eight autoregressive steps, although their FID is worse than at 32 steps.

\begin{figure}[!ht]
    \centering
    \includegraphics[width=0.48\textwidth]{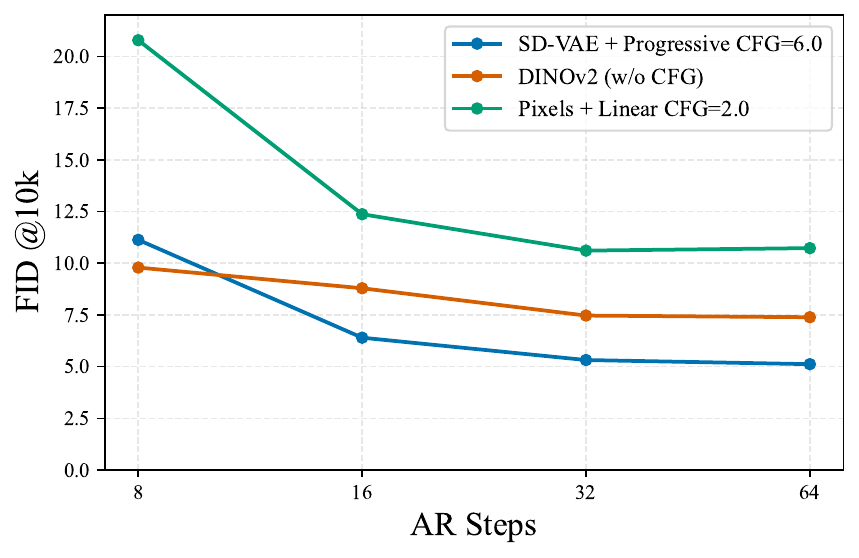} 
    \caption{
        \textbf{Effect of the number of autoregressive steps on FID@10k.}
        Each representation uses its final guidance configuration. With a sequence length of $N=256$, 32 autoregressive steps generate an average of eight new tokens per step.
    }
    \label{fig:comparison_ar_steps}
\end{figure}

The broadly similar trend indicates that all three representations support parallel masked generation. Their larger differences emerge in training efficiency, local denoising, and the number of flow-integration steps rather than in the overall number of autoregressive updates required for convergence.

\subsection{ODE Integration and One-Step Generation}
\begin{figure}[t]
     \centering
     \begin{subfigure}[t]{0.48\textwidth}
         \centering
         \includegraphics[width=\textwidth]{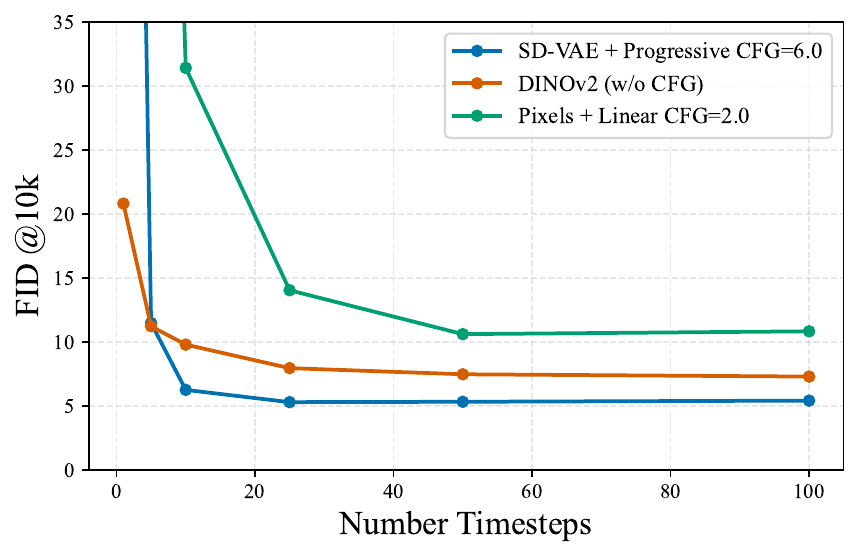}
         \caption{FID@10k vs. timesteps $t$}
     \end{subfigure}
     \hfill
     \begin{subfigure}[t]{0.48\textwidth}
         \centering
         \includegraphics[width=\textwidth]{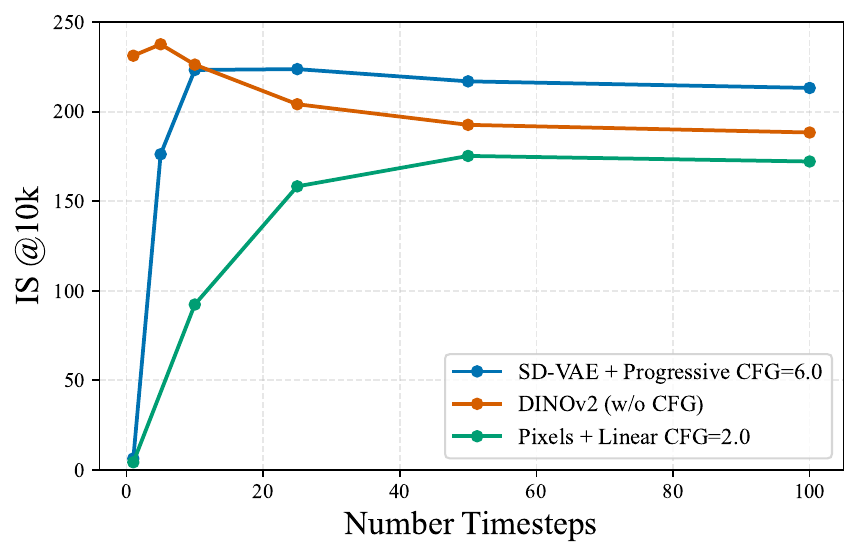}
         \caption{IS@10k vs. timesteps $t$}
     \end{subfigure}
     \caption{
        \textbf{Effect of the number of Euler integration steps per generated token.}
        We report FID@10k (a) and Inception Score@10k (b). SD-VAE uses progressive guidance with $w=6$, DINOv2 is unguided, and pixels use linear guidance with $w=2$.
     }
     \label{fig:comparison_sampling_t}
\end{figure}

We next vary the number of Euler integration steps used by the local rectified-flow sampler. Across all three representations, FID approaches convergence once the number of ODE steps exceeds approximately 25.

The representations differ sharply at the extreme of a single Euler step. DINOv2 obtains an FID@10k of 20.82 and an Inception Score of 231.31. Its generated images retain recognizable objects, sharp boundaries, and substantial class-specific structure.

Under the same single-step evaluation, SD-VAE obtains an FID@10k of 139.80 and an Inception Score of 6.30. Pixel space obtains an FID@10k of 190.05 and an Inception Score of 4.21. Qualitatively, the SD-VAE model produces only coarse outlines, while the DINOv2 model retains much clearer semantic and visual

\begin{table}[htb]
    \centering
    \caption{Comparing generation quality across representation spaces for a single timestep $t$ with the identical loss formulation. }
    \label{tab:single_timestep}
    
    \begin{tabular}{l c c}
        \noalign{\hrule height 1pt}
        \rowcolor{gray!20} \textbf{Space} & \textbf{FID@10k} $\downarrow$ & \textbf{IS@10k} $\uparrow$ \\
        \hline
        SD-VAE & \cellcolor{red!15} 139.80 & \cellcolor{red!15} 6.30 \\
        DINOv2-B & \cellcolor{green!15} 20.82 & \cellcolor{green!15} 231.31 \\
        Pixels         & \cellcolor{red!15} 190.05 & \cellcolor{red!15} 4.21 \\
        \noalign{\hrule height 1pt}
        \noalign{\vskip 0.3em}
    \end{tabular}
    
\end{table}

The result shows that DINOv2 supports a much more effective single-step approximation of the learned conditional transport. It does not establish that the underlying flow trajectory is globally straight or intrinsically simpler, but it demonstrates a large representation-dependent difference under an identical first-order sampling procedure.

In Fig.~\ref{fig:comparison_sampling_t_samples}, we provide a qualitative comparison of examples generated using a single timestep for both SD-VAE and DINOv2. While the DINOv2 model retains sharp structures and clear semantic definition, the SD-VAE baseline constructs only a coarse outline of the target object.

\begin{figure}[!htb]
     \centering
     \begin{subfigure}[t]{0.48\textwidth}
         \centering
         \includegraphics[width=\textwidth]{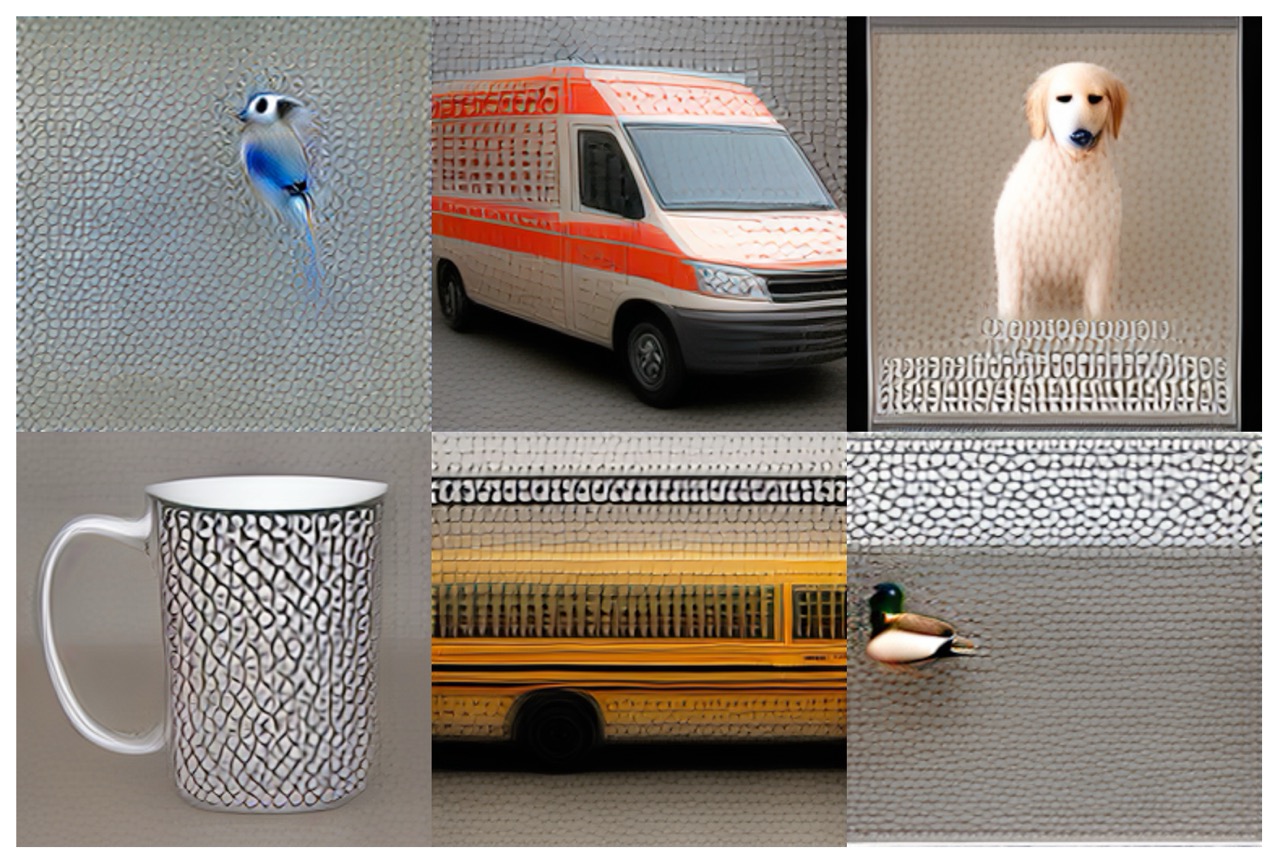}
         \caption{SD-VAE}
     \end{subfigure}
     \hfill
     \begin{subfigure}[t]{0.48\textwidth}
         \centering
         \includegraphics[width=\textwidth]{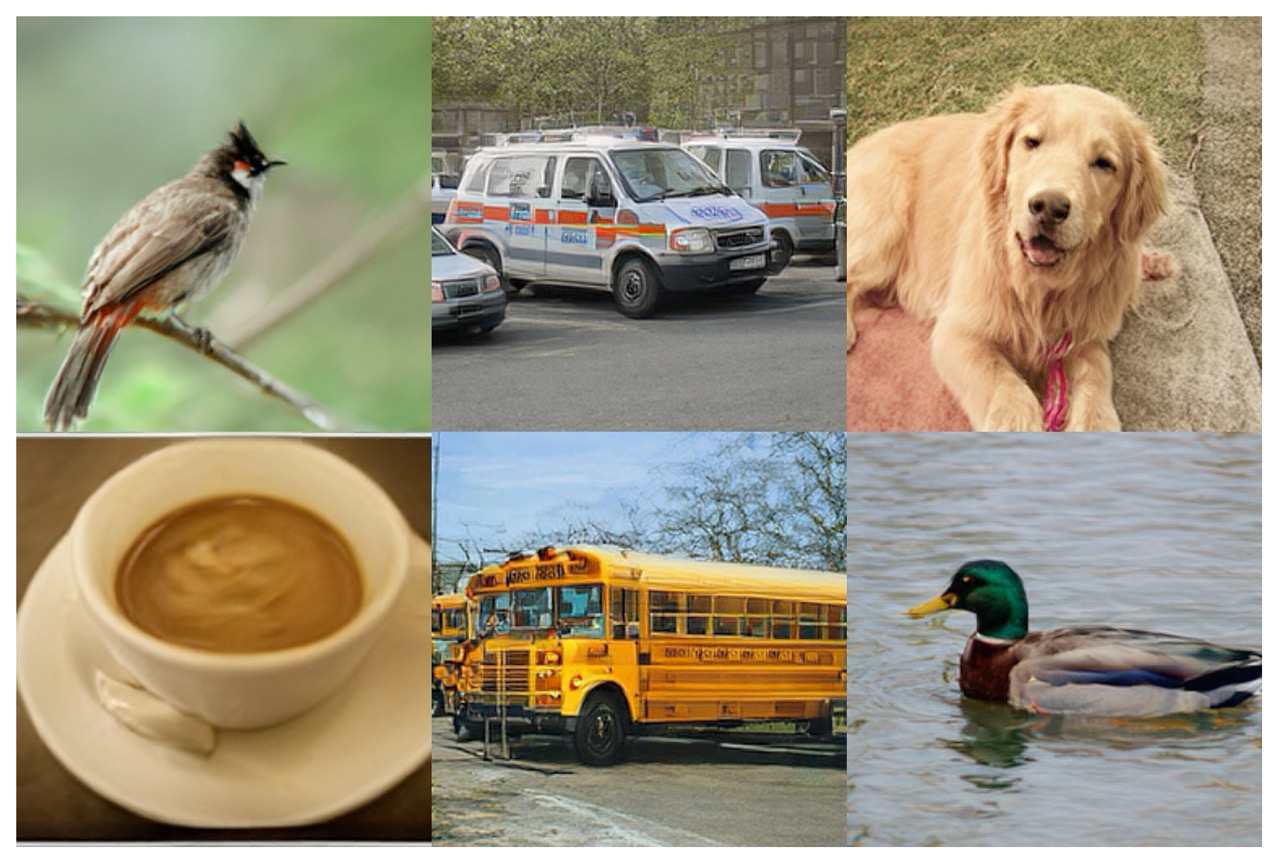}
         \caption{DINOv2-B}
     \end{subfigure}
     \caption{
        \textbf{Samples generated with a single Euler integration step.}
        DINOv2 retains recognizable class structure and comparatively sharp boundaries, whereas SD-VAE produces only coarse and degraded object outlines.
     }
     \label{fig:comparison_sampling_t_samples}
\end{figure}

\subsection{Autoregressive Generation Dynamics}
\begin{figure}[!htb]
     \centering
     \begin{subfigure}[b]{\textwidth}
         \centering
         \includegraphics[width=0.98\textwidth]{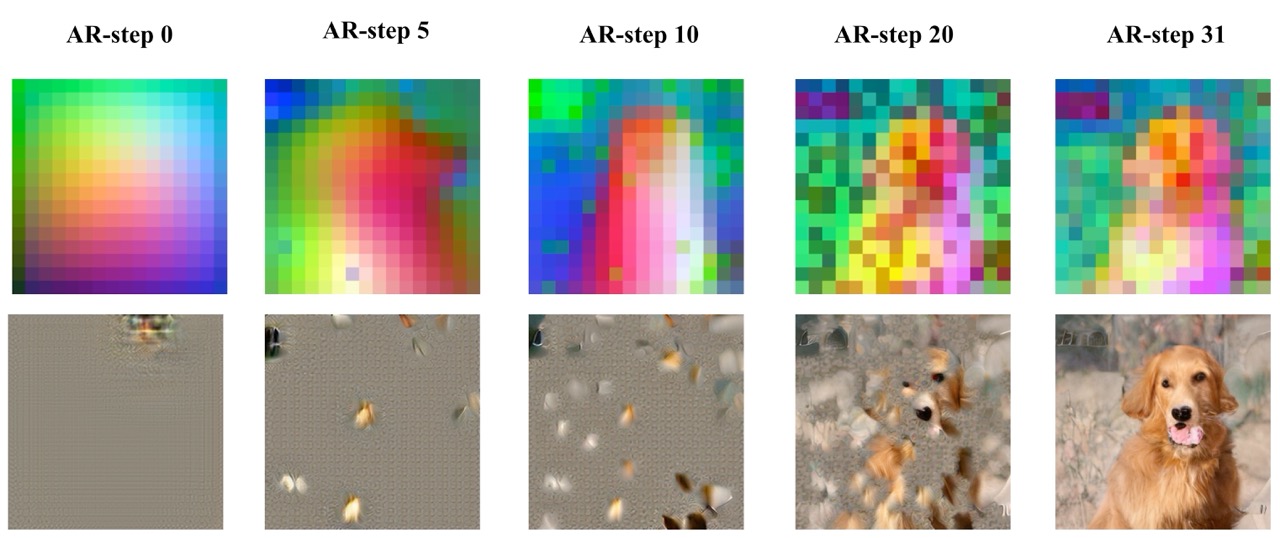}
         \caption{Autoregressive steps in SD-VAE latent space.}
         \label{fig:vae_ar_steps}
     \end{subfigure}
     \vspace{1em}
     \begin{subfigure}[b]{\textwidth}
         \centering
         \includegraphics[width=0.98\textwidth]{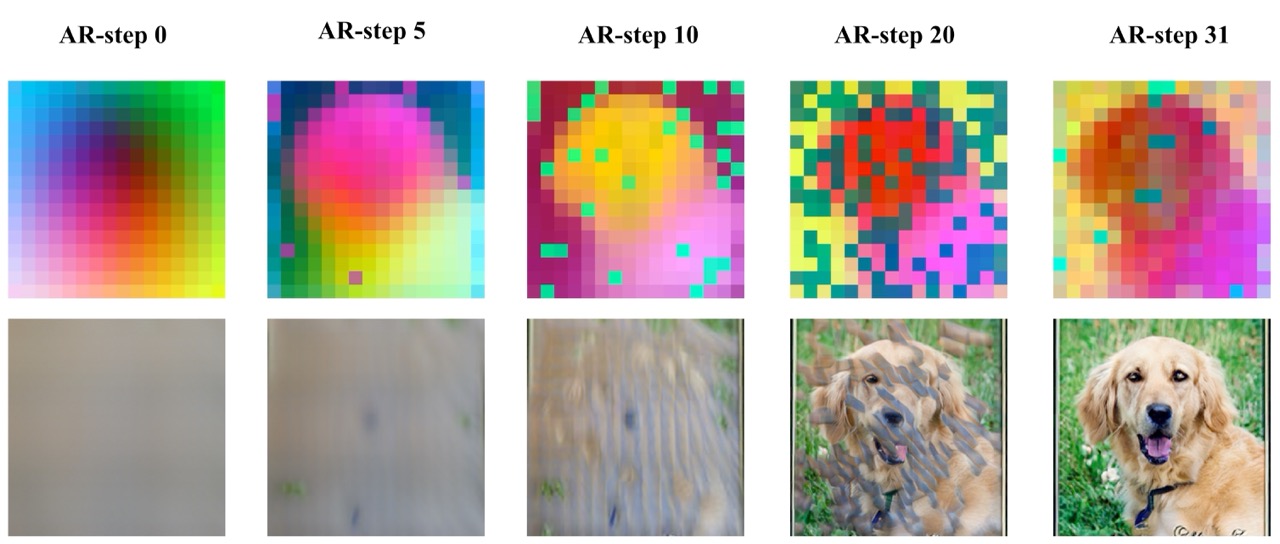}
         \caption{Autoregressive steps in DINOv2 latent space.}
         \label{fig:dinov2_ar_steps}
     \end{subfigure}
    \vspace{1em}
     \begin{subfigure}[b]{\textwidth}
         \centering
         \includegraphics[width=0.98\textwidth]{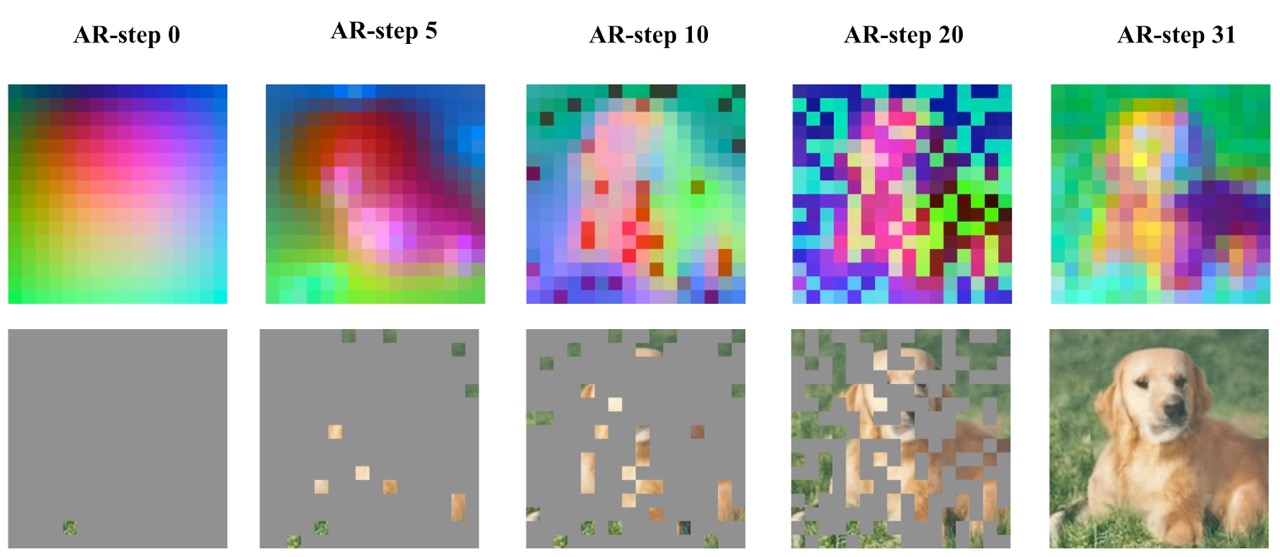}
         \caption{Autoregressive steps in pixel space.}
         \label{fig:pixel_ar_steps}
     \end{subfigure}
     \caption{
        \textbf{Autoregressive generation dynamics at steps 0, 5, 10, 20, and 31.}
        For each representation, the upper row shows the PCA-colored transformer context $z$, and the lower row shows the corresponding decoded partial image after the active tokens have been denoised.
     }
     \label{fig:visualize_ar_steps}
\end{figure}
We visualize the generation process at autoregressive steps 0, 5, 10, 20, and 31 in Fig.~\ref{fig:visualize_ar_steps}. For each representation, we show the PCA-colored transformer context z together with the decoded image after newly activated tokens have been denoised.

Across all spaces, the contextual features at already visible positions exhibit structural changes over the generation sequence. As in the masking analysis, these positions are not directly supervised by the flow loss. The generation itself is unaffected because the local denoiser only uses the contextual vectors for currently masked positions.

DINOv2 establishes a recognizable global semantic context after only a small number of tokens have been generated. Newly generated DINOv2 tokens also influence the decoded appearance of neighboring regions, reflecting the broader spatial footprint of the representation. SD-VAE exhibits an intermediate degree of interaction. Pixel tokens remain strictly localized, so newly generated patches do not directly alter neighboring patches in the decoded image.

These dynamics help explain why identical masking ratios produce different conditional problems. The visible-token count is the same, but the information carried by each token and its spatial footprint vary across representations.

\section{More on ``What Does Not Predict Generative Behavior?''}

We visualize the global organization of each target space using t-SNE in Fig.~\ref{fig:tsne}. We sample 20 ImageNet classes with 200 images per class, encode every image, average its spatial tokens, and project the resulting global features into two dimensions.

\begin{figure}[t]
     \centering
     \begin{subfigure}[t]{0.48\textwidth}
         \centering
         \includegraphics[width=\textwidth]{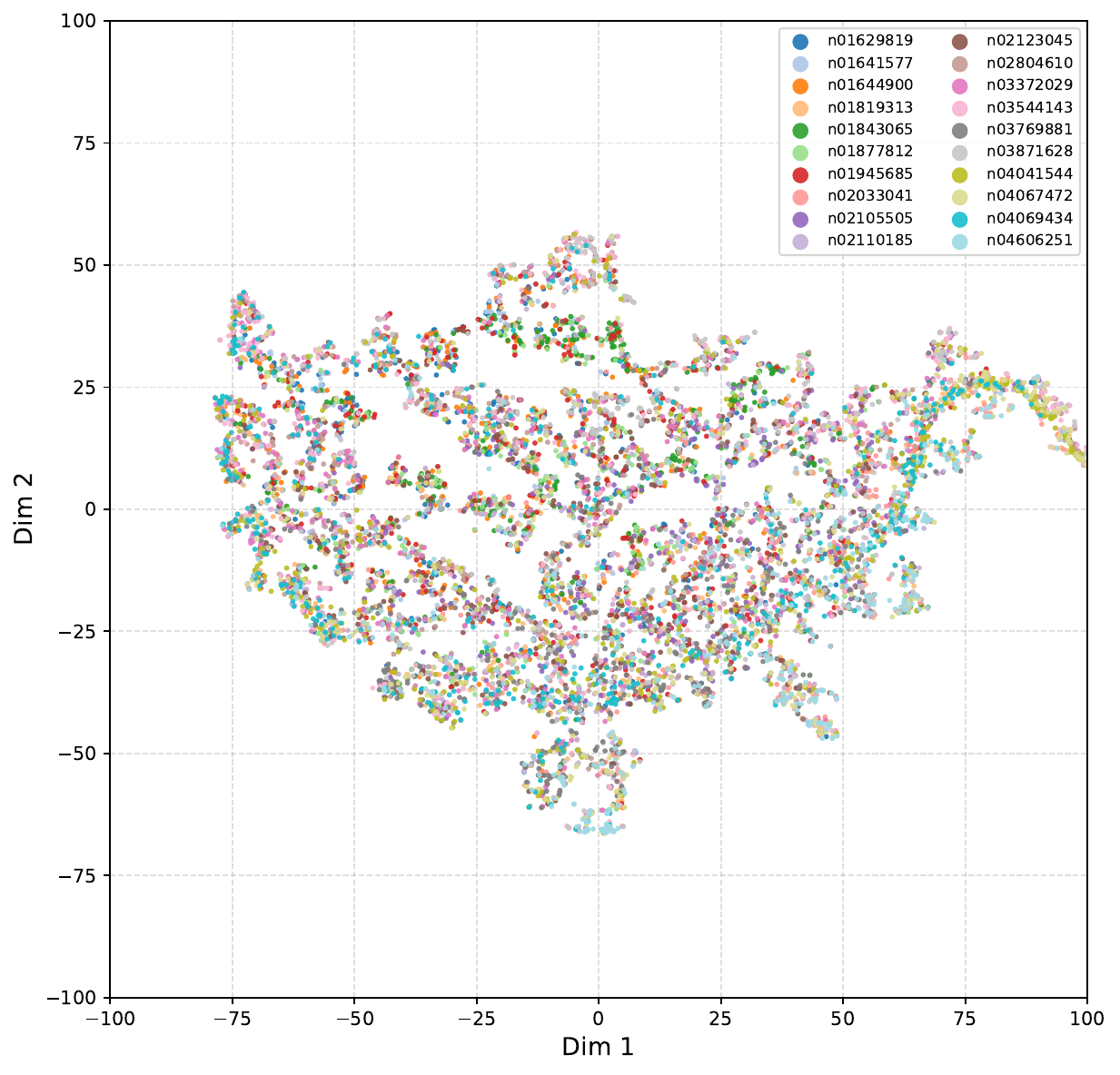}
         \caption{Pixel Space ($\text{dim}=768$)}
         \label{fig:tsne_pixel}
     \end{subfigure}
     \hfill
     \begin{subfigure}[t]{0.48\textwidth}
         \centering
         \includegraphics[width=\textwidth]{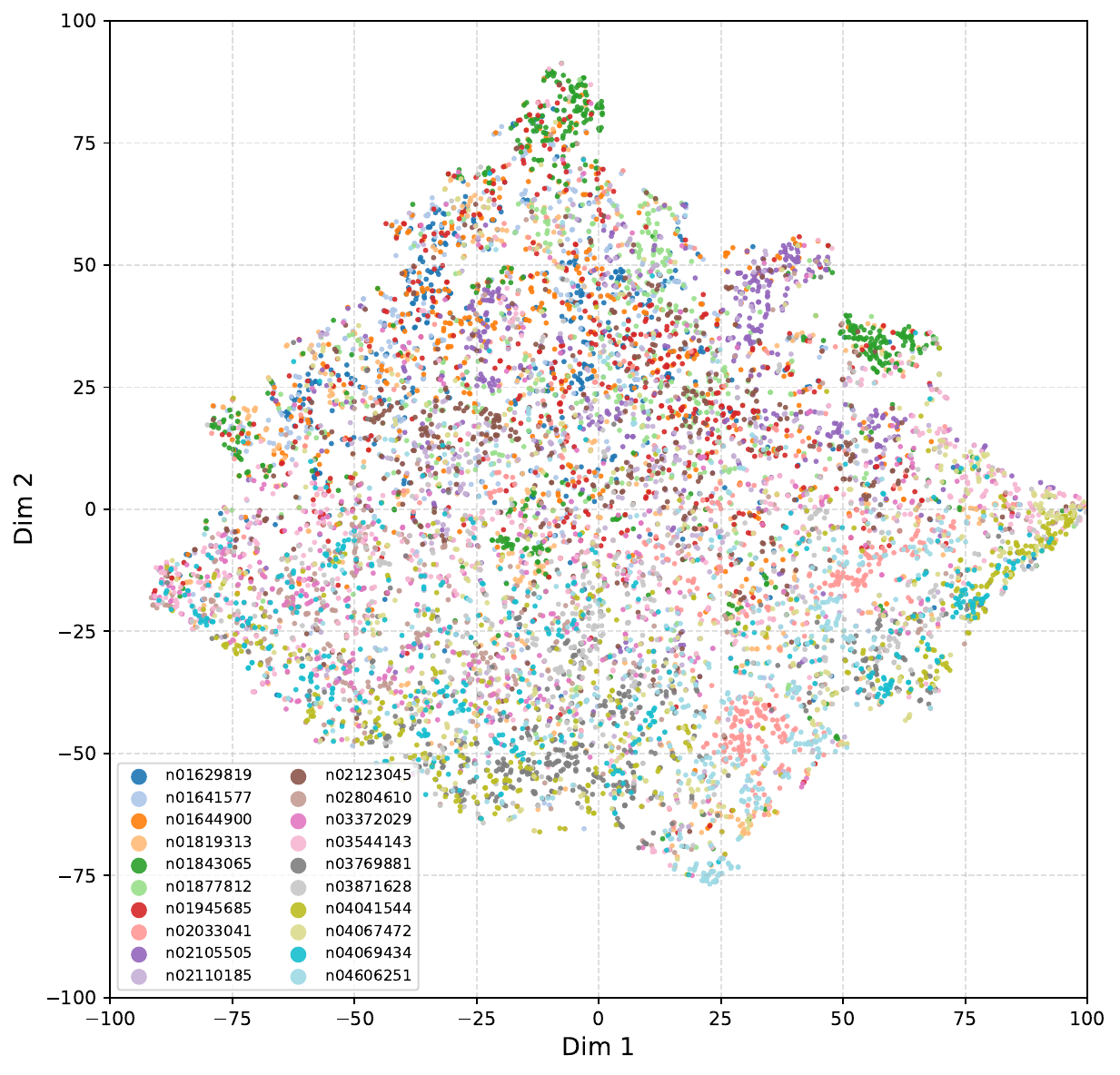}
         \caption{SD-VAE (KL-16) ($\text{dim}=16$)}
         \label{fig:tsne_vae}
     \end{subfigure}
     \vspace{1.5em} 
     \begin{subfigure}[t]{0.48\textwidth}
         \centering
         \includegraphics[width=\textwidth]{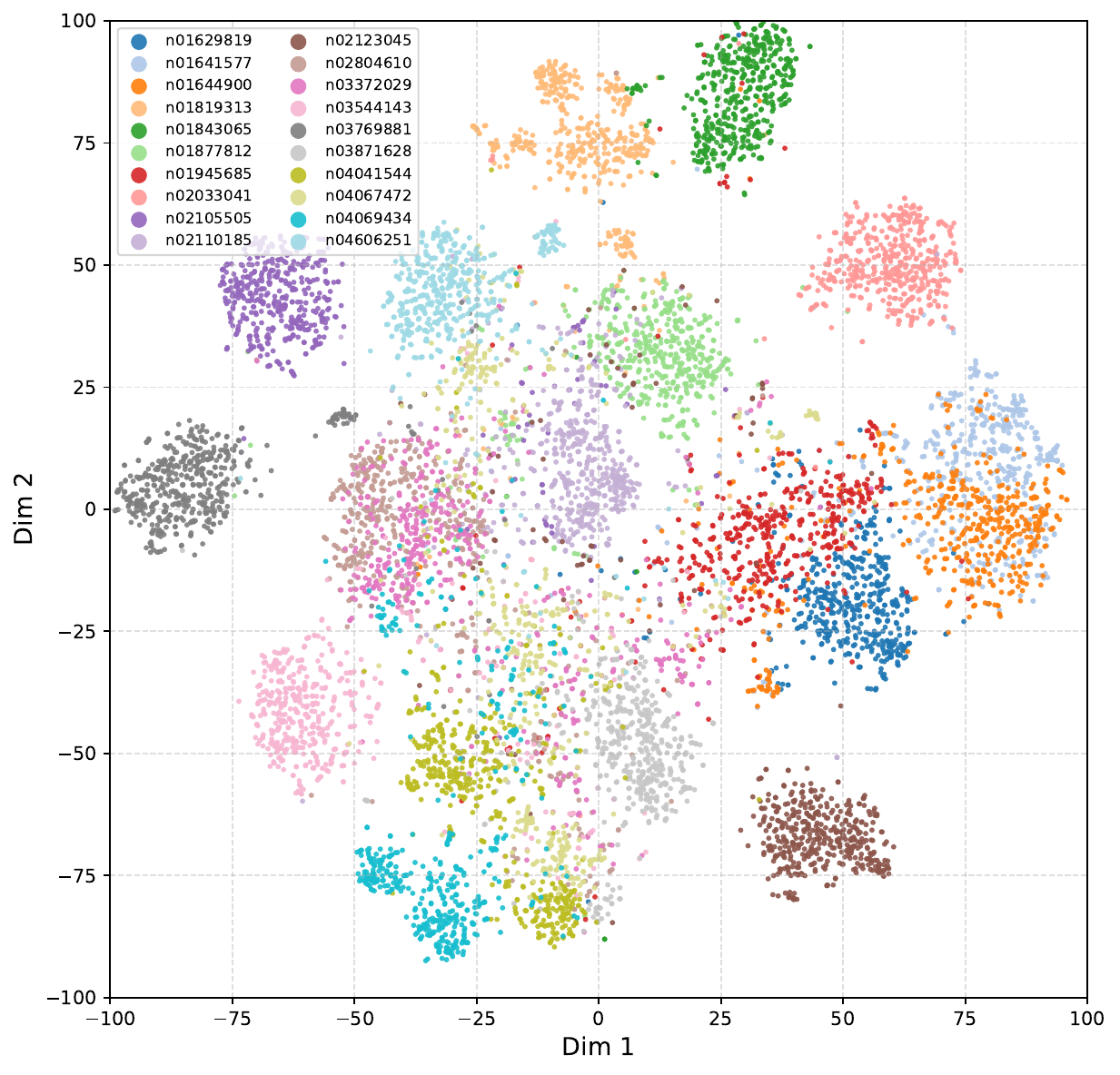}
         \caption{MAE-B ($\text{dim}=768$)}
         \label{fig:tsne_mae}
     \end{subfigure}
     \hfill
     \begin{subfigure}[t]{0.48\textwidth}
         \centering
         \includegraphics[width=\textwidth]{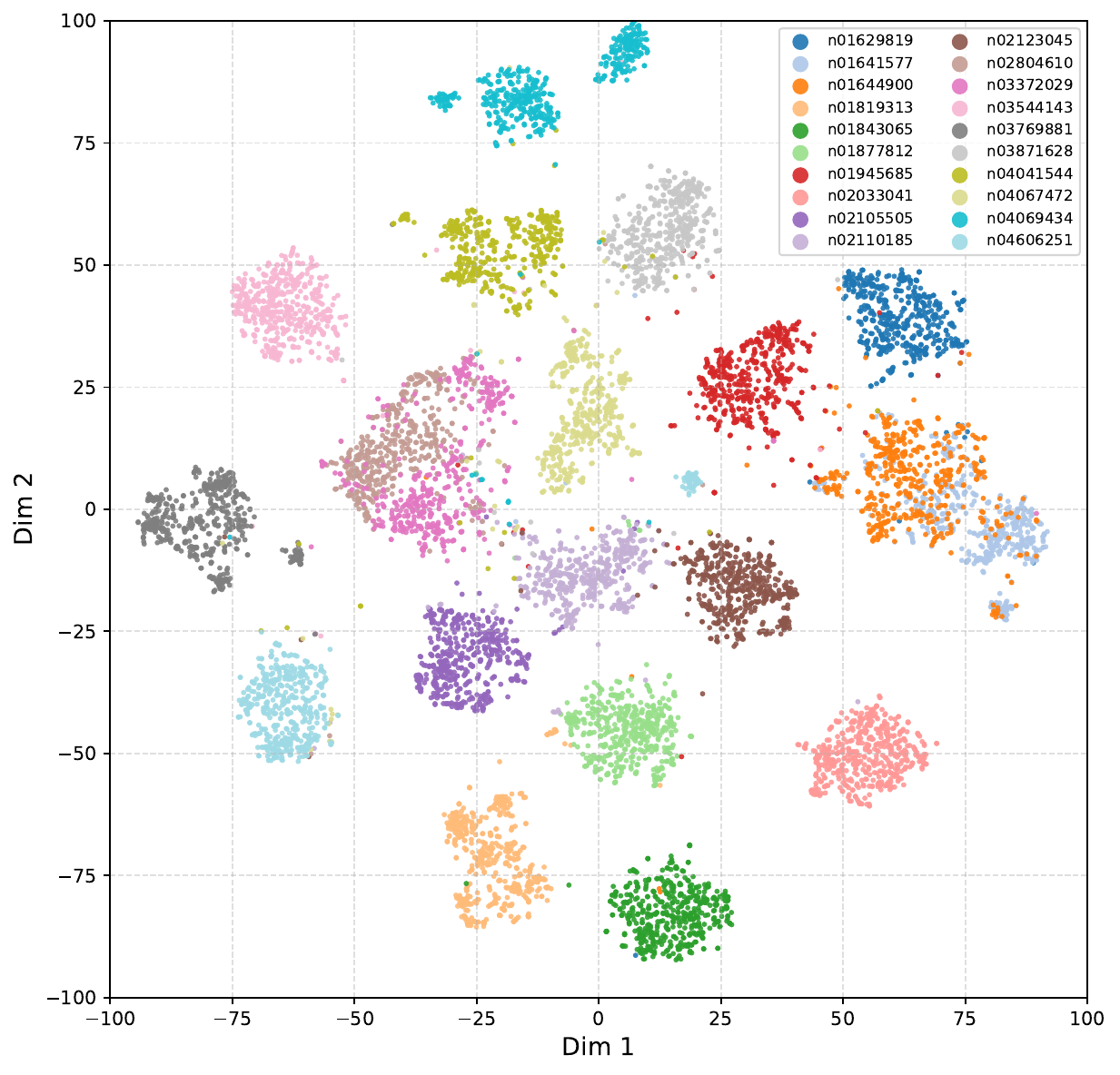}
         \caption{DINOv2-B ($\text{dim}=768$)}
         \label{fig:tsne_dinov2}
     \end{subfigure}
     \caption{
        \textbf{Qualitative visualization of global representation structure.}
        We spatially average the tokens of 200 images from each of 20 ImageNet classes and project them with t-SNE. DINOv2-B and MAE-B exhibit clearer class separation than SD-VAE and pixels. The visualization is qualitative and does not characterize the complete token geometry.
     }
     \label{fig:tsne}
\end{figure}

\subsection{Reconstruction Fidelity: Qualitative Examples}
In Fig. \ref{fig:reconstructed_images} we show the reconstructed examples on ImageNet after we applied the autoencoders. We can clearly see that both scales of DINOv2 lack behind SD-VAE and MAE-B, while MAE-B shows the overall best reconstructions.

\begin{figure}[!htb]
    \centering
    \includegraphics[width=\textwidth]{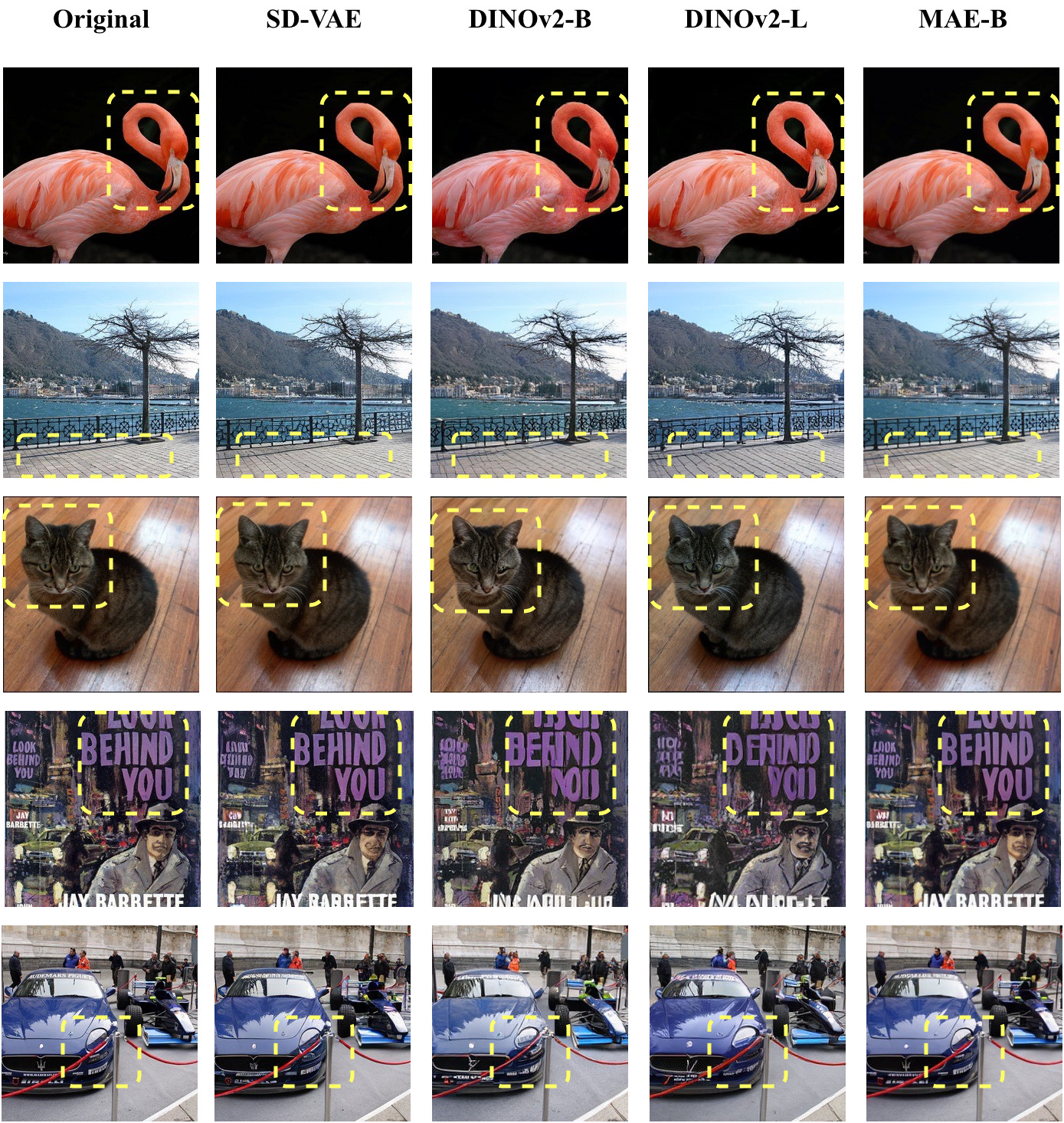} 
    \caption{
        \textbf{Reconstructions produced by the frozen SD-VAE, DINOv2-B, DINOv2-L, and MAE-B encoder–decoder pairs.}
        MAE preserves the most image detail and obtains the best reconstruction metrics, whereas DINOv2 shows larger perceptual reconstruction errors despite yielding substantially better masked-generation results.
    }
    \label{fig:reconstructed_images}
\end{figure}

\subsection{Semantic Organization}
Average-pooled pixel patches primarily retain global color statistics and do not produce distinct category clusters. SD-VAE features form a compact central distribution without clear class separation. This is consistent with an encoder optimized for reconstruction and regularized toward a Gaussian prior rather than for class-level organization.

Both DINOv2 and MAE produce clearly separated class clusters. Despite their different pretraining objectives, each representation contains visible category-level semantic structure under the selected projection.

The generative results nevertheless differ considerably between the two spaces. Clear semantic clustering is therefore not sufficient to predict whether a target space will be modeled effectively by the masked generator.

The t-SNE visualization is qualitative and depends on spatial average pooling and dimensionality reduction. We use it only to establish that both DINOv2 and MAE exhibit visible semantic organization, not as a quantitative measurement of their full token geometry.

\subsection{Compression and Dimensionality}
Compression also fails to provide a monotonic explanation. The 16-dimensional SD-VAE representation is easier to optimize than 768-dimensional pixels, consistent with a dimensionality-based account. However, 768-dimensional DINOv2 features are easier to optimize than both of them.

Likewise, the two 768-dimensional spaces of DINOv2 and pixels occupy opposite ends of the optimization ranking. Token dimensionality therefore interacts strongly with representation structure and model design.

The results do not imply that dimensionality is irrelevant. Both DINOv2 and pixels benefit strongly from a wider local denoiser, whereas the SD-VAE model operates with a smaller denoiser expansion ratio. Instead, dimensionality appears to influence where capacity is needed without independently determining whether the resulting distribution is easy to learn.

\subsection{Decoded Space Visualization}
In Fig.~\ref{fig:comparing_tokenizers} we visualize each representation space alongside its corresponding decoded spatial image after a masking schedule has been applied directly within that space. This visualization allows us to systematically understand how spatial masking effects the preserved information across different semantic configurations. We see that in SD-VAE  and DINOv2 a single token effects also neighboring tokens while MAE does not. 

\begin{figure}[!htb]
    \centering
    \includegraphics[width=\textwidth]{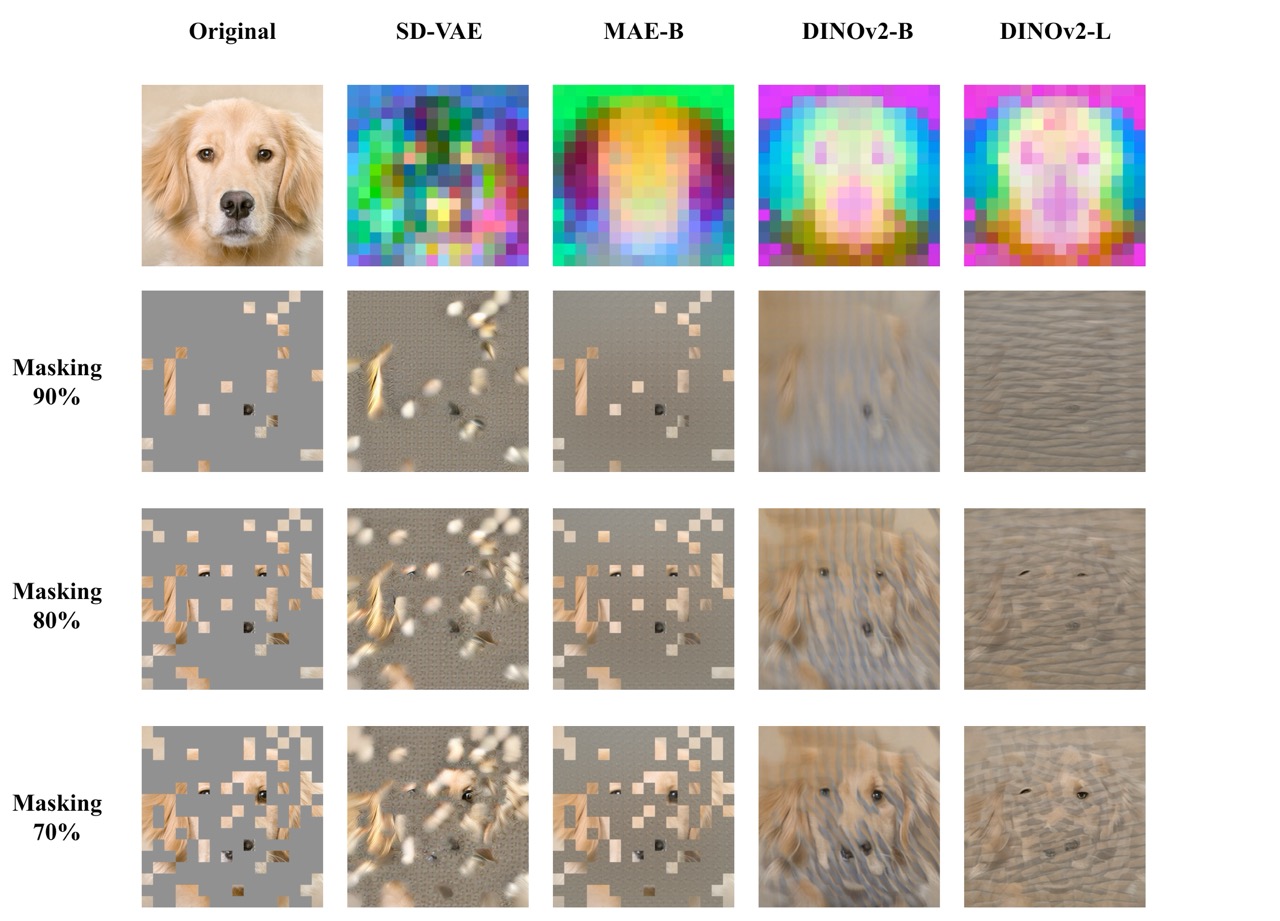} 
    \caption{
        \textbf{Effect of spatial token masking in different representation spaces.}
        Rows show 90\%, 80\%, and 70\% masking. For each latent representation, the remaining tokens are decoded into pixel space. Masking one SD-VAE or DINOv2 token affects a broader spatial region, whereas the effect is more localized in MAE space.
    }
    \label{fig:comparing_tokenizers}
\end{figure}

\section{Discussion}

\subsection{Target Representations Induce Different Optimization Regimes}

Our experiments show that a target representation cannot be treated as an interchangeable output interface. SD-VAE, DINOv2, MAE, and pixels require different combinations of timestep sampling, prediction parameterization, masking, local denoiser design, and classifier-free guidance.

DINOv2 converges most rapidly under the shared training budget and retains useful generation with one Euler step. It achieves this with a relatively shallow contextual encoder, but benefits substantially from a wider local denoiser and direct context fusion.

Pixels are much slower to optimize and require a sequence of coordinated changes: clean-data prediction, a less aggressive mask distribution, a bottleneck embedding, a wider denoiser, direct context fusion, and linear classifier-free guidance. These changes improve pixel generation from an unguided FID above 100 in the initial configuration to a guided FID of 8.70.

SD-VAE occupies an intermediate regime. Its compressed token dimension permits a smaller local denoiser, but unguided optimization is slower than DINOv2. It benefits strongly from progressive guidance and ultimately obtains the best guided FID.

MAE demonstrates that representation-autoencoder features do not behave uniformly. Despite strong reconstruction and semantic clustering, MAE generation remains worse than DINOv2 and depends on linear guidance.

The evidence therefore supports a representation-dependent allocation of modeling difficulty. The experiments do not yield a scalar measure that predicts this allocation, nor do they establish one property as its unique cause. Instead, they identify where the consequences become visible: contextual reconstruction under masking, local denoiser sensitivity, flow-integration efficiency, and inference-time guidance.

\subsection{Reconstruction and Generative Learnability Are Distinct}

Reconstruction fidelity determines whether a frozen decoder can map generated tokens into detailed images. It does not determine whether the token distribution itself is easy to learn.

MAE provides the best reconstructions but substantially worse generation than DINOv2. DINOv2 reconstructs less faithfully according to LPIPS but is learned much more efficiently. SD-VAE achieves strong guided generation despite lacking the visible class clustering observed in DINOv2 and MAE.

A useful target representation must therefore satisfy at least two distinct requirements: it must preserve sufficient image information for decoding, and it must expose a distribution that the generator can model effectively. Improvements in one requirement do not automatically improve the other.

\subsection{Context and Local Transport Must Be Considered Jointly}

Masked generation separates inter-token context construction from per-token denoising. The representation influences both.

DINOv2 tokens retain coherent global context under severe masking and influence broader spatial regions. This can make a small number of visible tokens informative early in autoregressive generation. At the same time, each missing token is 768-dimensional and benefits from a strong local model.

Pixel tokens are equally high-dimensional but strictly localized. They provide less information about neighboring locations, and the model benefits from a less aggressive masking distribution. Their local denoising also benefits from clean-data prediction, a low-rank bottleneck, and increased capacity.

The resulting behavior cannot be summarized as “semantic representations are easy” or “high-dimensional representations are hard.” DINOv2 and pixels demonstrate that contextual structure and local target complexity can vary independently even when token dimensionality is identical.

\subsection{Distributional Control Is Representation-Dependent}

Classifier-free guidance has qualitatively different effects across the spaces. SD-VAE benefits from progressive guidance, while pixels and MAE require linear guidance from the early generation steps. DINOv2 does not improve with either strategy.

The representation also changes the precision-recall balance. DINOv2 achieves the highest precision but the lowest recall. SD-VAE provides broader coverage and reaches the best guided FID. Pixels remain less precise than both latent models but maintain higher unguided recall than DINOv2.

Optimization efficiency should therefore not be equated with complete distribution modeling. A representation can converge rapidly toward high-fidelity samples while covering a smaller portion of the data distribution.

\section{Empirical Principles}
Across the target spaces and model configurations studied in this work, 
our experiments suggest five empirical principles:

\begin{itemize}

    \item \textbf{Target dimensionality alone does not determine generative difficulty.}
    DINOv2 and pixels both use 768-dimensional tokens, yet exhibit the fastest
    and slowest unguided optimization, respectively, while the 16-dimensional
    SD-VAE representation lies between them.

    \item \textbf{Reconstruction fidelity is a constraint, but not a sufficient
    predictor of generative learnability.}
    MAE achieves the strongest reconstruction metrics, but yields substantially
    worse generation than DINOv2.

    \item \textbf{Visible semantic organization is insufficient to predict
    generative learnability.}
    DINOv2 and MAE both form clearly separated class clusters under our t-SNE
    visualization, yet differ markedly in generation quality and guidance
    dependence.

    \item \textbf{Representation choice redistributes modeling demands across
    model components.}
    Our final DINOv2 configuration uses a two-layer context encoder but benefits
    strongly from a wider local denoiser and direct context fusion, whereas pixel
    generation benefits from a different combination of prediction,
    masking, bottleneck, denoising, and guidance choices.

    \item \textbf{Optimization efficiency and distribution coverage are distinct.}
    DINOv2 converges fastest and achieves the highest precision among the
    unguided models, but also exhibits the lowest recall.

\end{itemize}

\section{Limitations}

\noindent\textbf{Frozen decoder dependence.}
Generation in SD-VAE, DINOv2, and MAE space is constrained by a pretrained decoder. In particular, the DINOv2 representation-autoencoder has worse LPIPS reconstruction than SD-VAE and MAE. The RAE decoders are trained with injected feature noise to improve robustness to off-manifold inputs, but the distribution of these training perturbations may not match the errors produced by our masked generator.

\noindent\textbf{Representation-specific configurations.}
The models share a common masked autoregressive flow factorization, spatial token layout, dataset, training duration, and evaluation protocol. Their final configurations are not identical. We adapt timestep sampling, masking, prediction parameterization, transformer depth, and denoiser design within each space. The final cross-representation ranking therefore compares the resulting representation-specific systems rather than varying only the frozen encoder under an otherwise identical model.

\noindent\textbf{Incomplete hyperparameter search.}
We evaluate a targeted set of established configurations rather than performing an exhaustive optimization for every representation. Additional combinations of masking, timestep sampling, model capacity, and guidance may further improve each space.

\noindent\textbf{Model scale.}
Our models range from approximately 165M to 222M parameters in the principal comparison, with one 255M DINOv2 capacity ablation. We do not evaluate large or huge model scales. The observed allocation of difficulty may change as both the context transformer and local denoiser are scaled.

\noindent\textbf{Dataset and resolution.}
All experiments use class-conditional ImageNet-1k generation at 256×256. We do not evaluate text conditioning, higher resolutions, or datasets with different structural and semantic distributions.

\noindent\textbf{Qualitative context analysis.}
Our masking and autoregressive visualizations show clear representation-dependent differences, but they do not directly quantify conditional entropy or masked-token predictability. We therefore use them as evidence about the behavior of the trained context model rather than as a complete causal explanation of convergence.

\noindent\textbf{Diversity in DINOv2 space.}
DINOv2’s low recall remains unresolved. We do not evaluate alternative guidance methods such as AutoGuidance or representation alignment, and we cannot determine whether the reduced coverage originates from the representation, decoder, model scale, or training objective.

\endgroup

\clearpage
\section{Qualitative Results}
\label{app:qualitative_results}

\begin{figure}[!htb]
    \centering
    \begin{subfigure}[b]{0.48\textwidth}
        \centering
        \includegraphics[width=\textwidth]{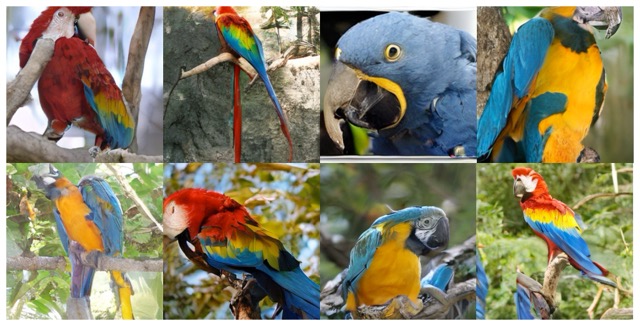}
        \caption{class 088: Parrot}
    \end{subfigure}
    \hfill
    \begin{subfigure}[b]{0.48\textwidth}
        \centering
        \includegraphics[width=\textwidth]{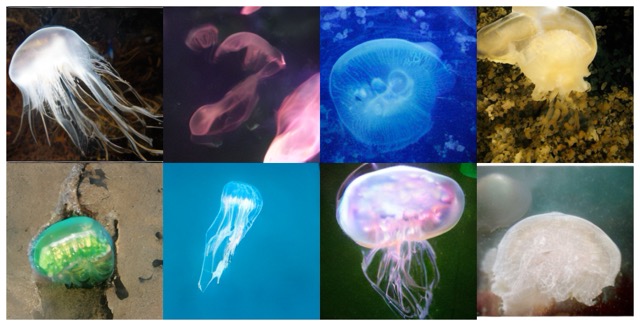}
        \caption{class 107: Jellyfish}
    \end{subfigure}
    \vspace{1.5em}
    \begin{subfigure}[b]{0.48\textwidth}
        \centering
        \includegraphics[width=\textwidth]{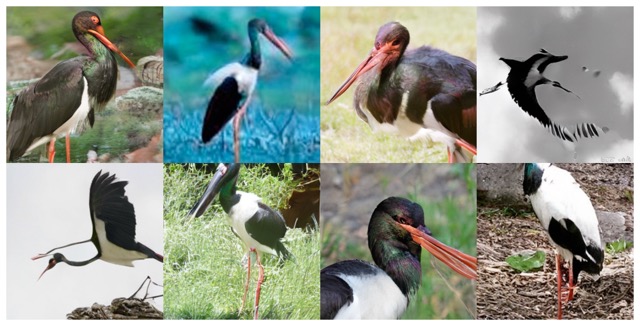}
        \caption{class 128: White stork}
    \end{subfigure}
    \hfill
    \begin{subfigure}[b]{0.48\textwidth}
        \centering
        \includegraphics[width=\textwidth]{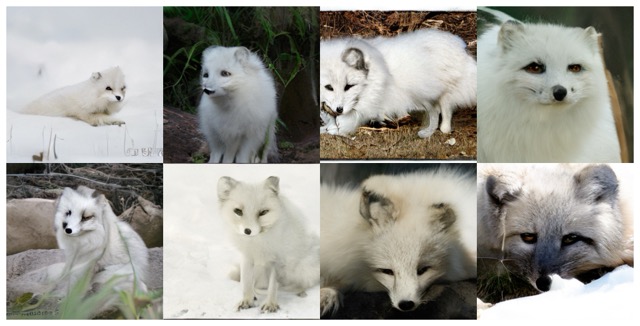}
        \caption{class 279: Arctic fox}
    \end{subfigure}
    \vspace{1.5em} 
    \begin{subfigure}[b]{0.48\textwidth}
        \centering
        \includegraphics[width=\textwidth]{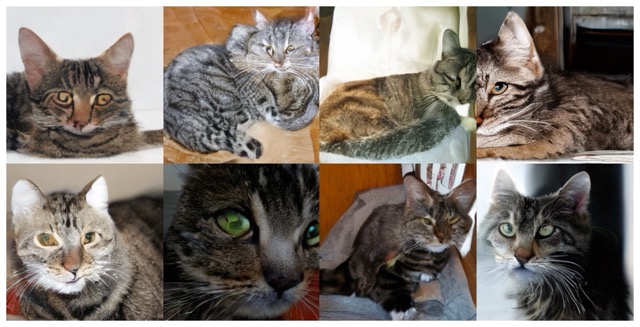}
        \caption{class 281: Tabby cat}
    \end{subfigure}
    \hfill
    \begin{subfigure}[b]{0.48\textwidth}
        \centering
        \includegraphics[width=\textwidth]{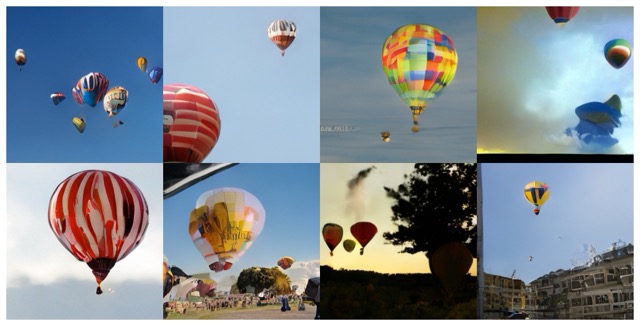}
        \caption{class 417: Balloon}
    \end{subfigure}
    \vspace{1.5em} 
    \begin{subfigure}[b]{0.48\textwidth}
        \centering
        \includegraphics[width=\textwidth]{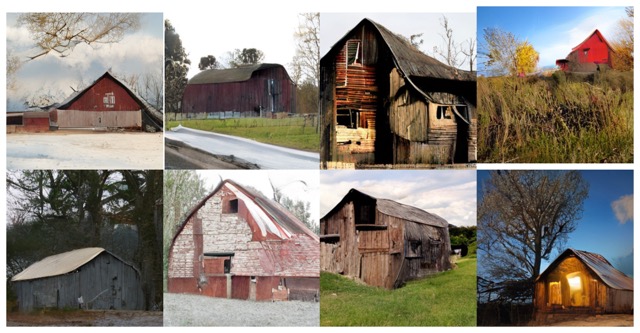}
        \caption{class 425: Barn}
    \end{subfigure}
    \hfill
    \begin{subfigure}[b]{0.48\textwidth}
        \centering
        \includegraphics[width=\textwidth]{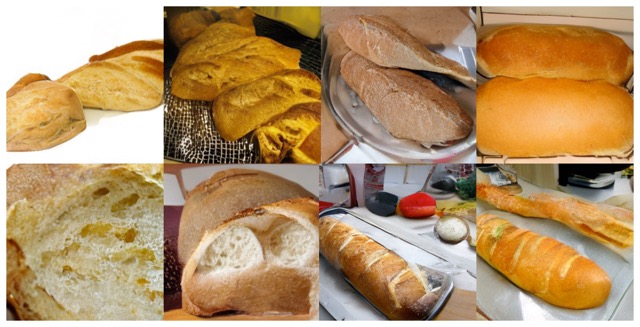}
        \caption{class 930: French loaf}
    \end{subfigure}
    \caption{\textbf{Additional class-conditional samples from the final SD-VAE model using progressive guidance with $w=6$.}}
    \label{fig:add_vae_samples}
\end{figure}

\begin{figure}[!htb]
    \centering
    \begin{subfigure}[b]{0.48\textwidth}
        \centering
        \includegraphics[width=\textwidth]{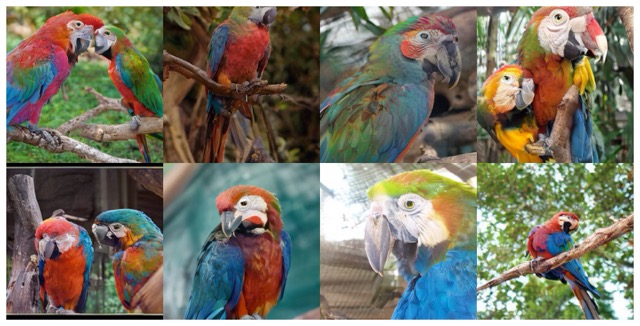}
        \caption{class 088: Parrot}
    \end{subfigure}
    \hfill
    \begin{subfigure}[b]{0.48\textwidth}
        \centering
        \includegraphics[width=\textwidth]{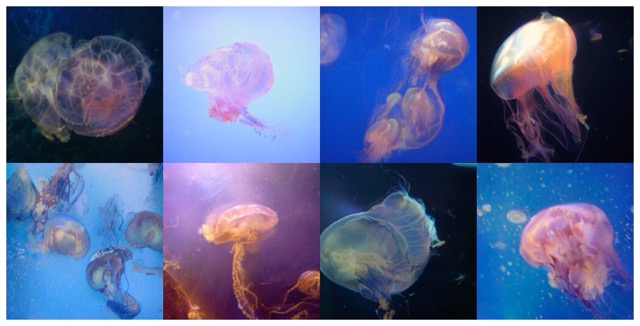}
        \caption{class 107: Jellyfish}
    \end{subfigure}
    \vspace{1.5em}
    \begin{subfigure}[b]{0.48\textwidth}
        \centering
        \includegraphics[width=\textwidth]{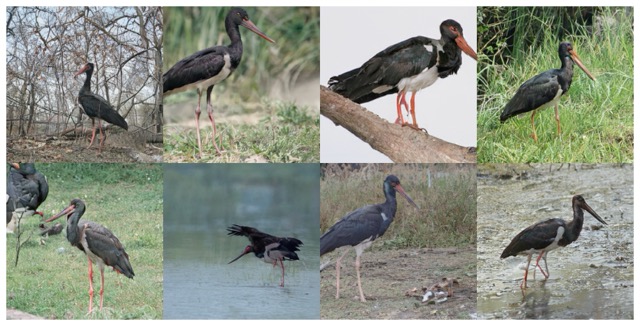}
        \caption{class 128: White stork}
    \end{subfigure}
    \hfill
    \begin{subfigure}[b]{0.48\textwidth}
        \centering
        \includegraphics[width=\textwidth]{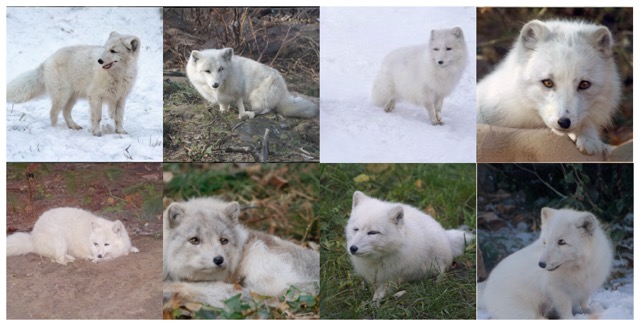}
        \caption{class 279: Arctic fox}
    \end{subfigure}
    \vspace{1.5em} 
    \begin{subfigure}[b]{0.48\textwidth}
        \centering
        \includegraphics[width=\textwidth]{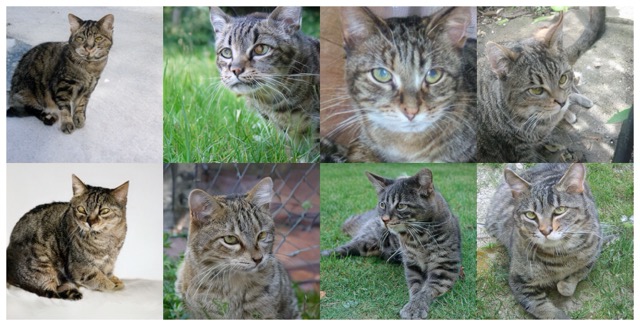}
        \caption{class 281: Tabby cat}
    \end{subfigure}
    \hfill
    \begin{subfigure}[b]{0.48\textwidth}
        \centering
        \includegraphics[width=\textwidth]{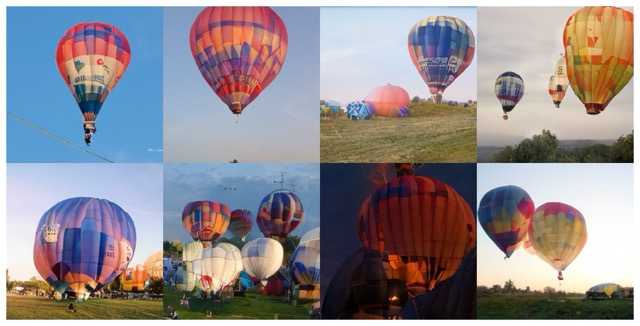}
        \caption{class 417: Balloon}
    \end{subfigure}
    \vspace{1.5em} 
    \begin{subfigure}[b]{0.48\textwidth}
        \centering
        \includegraphics[width=\textwidth]{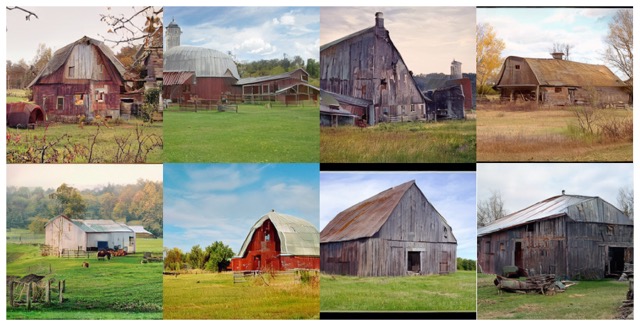}
        \caption{class 425: Barn}
    \end{subfigure}
    \hfill
    \begin{subfigure}[b]{0.48\textwidth}
        \centering
        \includegraphics[width=\textwidth]{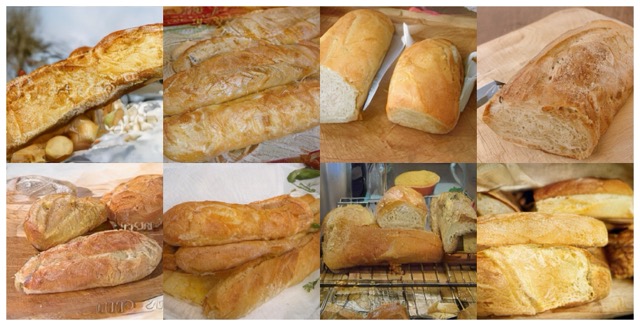}
        \caption{class 930: French loaf}
    \end{subfigure}
    \caption{\textbf{Additional class-conditional samples from the final unguided DINOv2-B model.}}
    \label{fig:add_dinov2_samples}
\end{figure}

\begin{figure}[!htb]
    \centering
    \begin{subfigure}[b]{0.48\textwidth}
        \centering
        \includegraphics[width=\textwidth]{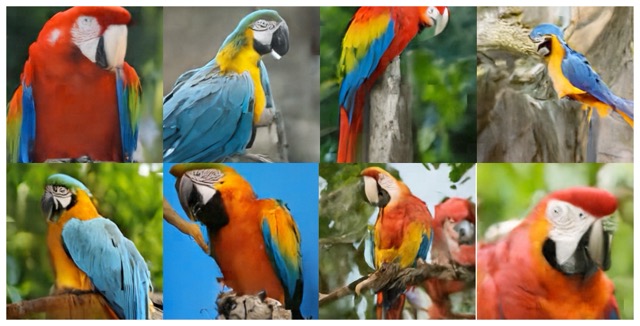}
        \caption{class 088: Parrot}
    \end{subfigure}
    \hfill
    \begin{subfigure}[b]{0.48\textwidth}
        \centering
        \includegraphics[width=\textwidth]{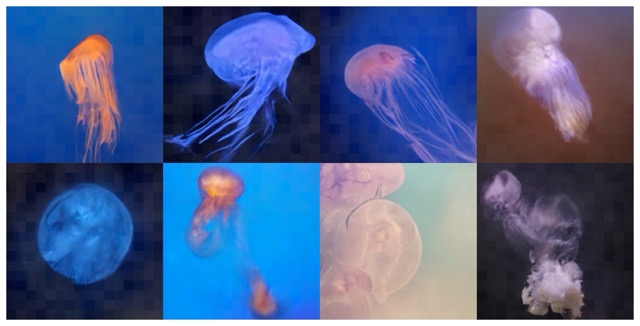}
        \caption{class 107: Jellyfish}
    \end{subfigure}
    \vspace{1.5em}
    \begin{subfigure}[b]{0.48\textwidth}
        \centering
        \includegraphics[width=\textwidth]{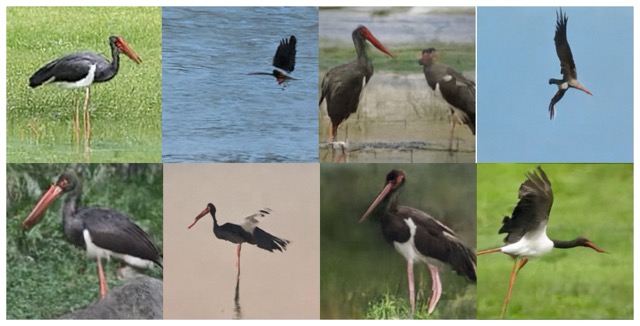}
        \caption{class 128: White stork}
    \end{subfigure}
    \hfill
    \begin{subfigure}[b]{0.48\textwidth}
        \centering
        \includegraphics[width=\textwidth]{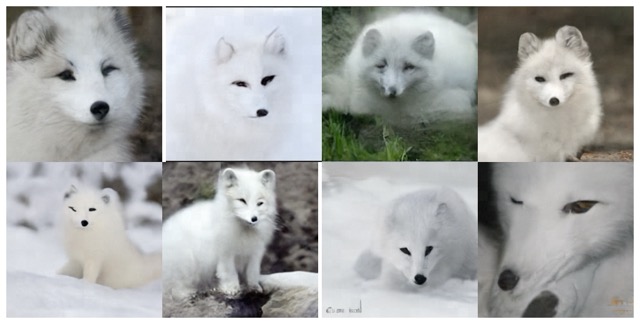}
        \caption{class 279: Arctic fox}
    \end{subfigure}
    \vspace{1.5em} 
    \begin{subfigure}[b]{0.48\textwidth}
        \centering
        \includegraphics[width=\textwidth]{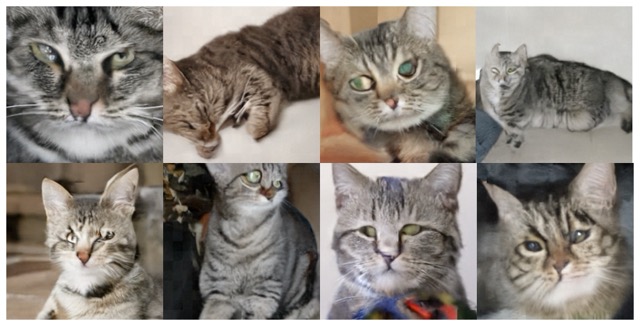}
        \caption{class 281: Tabby cat}
    \end{subfigure}
    \hfill
    \begin{subfigure}[b]{0.48\textwidth}
        \centering
        \includegraphics[width=\textwidth]{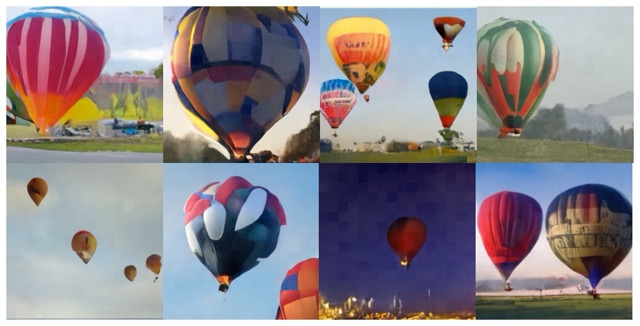}
        \caption{class 417: Balloon}
    \end{subfigure}
    \vspace{1.5em} 
    \begin{subfigure}[b]{0.48\textwidth}
        \centering
        \includegraphics[width=\textwidth]{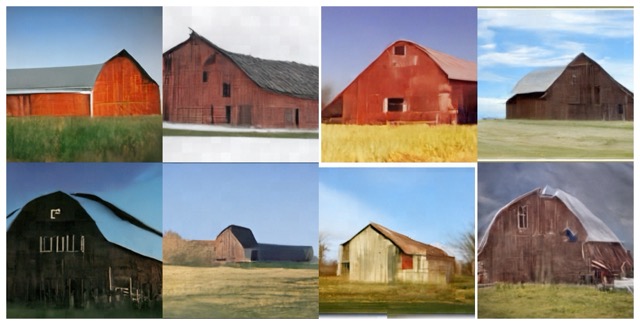}
        \caption{class 425: Barn}
    \end{subfigure}
    \hfill
    \begin{subfigure}[b]{0.48\textwidth}
        \centering
        \includegraphics[width=\textwidth]{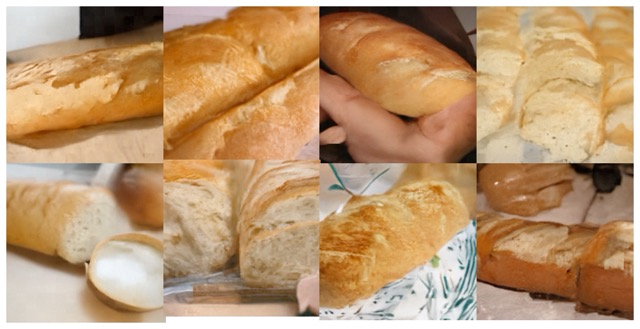}
        \caption{class 930: French loaf}
    \end{subfigure}
    \caption{\textbf{Additional class-conditional samples from the final pixel model using linear guidance with $w=2$.}}
    \label{fig:add_pixel_samples}
\end{figure}

\endgroup
\end{document}